\documentclass[11pt]{article}
\pdfoutput=1
\usepackage{authblk}

\def\pb{\pagebreak}   

\usepackage[dvipsnames]{xcolor}

\usepackage[framemethod=default]{mdframed}
\usepackage{caption}

\usepackage{bbm}
\usepackage{indentfirst}
\usepackage{bm, mathrsfs, graphics,float,amssymb,amsmath,subeqnarray,setspace,graphicx,amsthm,epstopdf,subfigure, enumerate, color}
\usepackage[utf8]{inputenc}
\usepackage[colorlinks,
linkcolor=red,
anchorcolor=blue,
citecolor=blue
]{hyperref}
\usepackage{natbib}
\usepackage{verbatim}

\usepackage{fullpage}
\numberwithin{equation}{section}

\newtheorem{lemma}{Lemma}
\newtheorem{theorem}{Theorem}
\newtheorem{proposition}{Proposition}

\newtheorem{remark}{Remark}

\ifx\assumption\undefined
\newtheorem{assumption}{Assumption}
\fi

\usepackage{multirow}
\usepackage{tablefootnote}
\usepackage{colortbl}
\usepackage{hhline}

\usepackage{algorithm}
\usepackage{algorithmic}

\usepackage[thinc]{esdiff}

\usepackage{mathtools}

\allowdisplaybreaks  

\begin{document}
\title{Mean-Field Analysis of Two-Layer Neural Networks: \\ Global Optimality with Linear Convergence Rates}

\author[$\ddag$ $\dag$]{Jingwei Zhang}

\author[$\ddag$]{
Xunpeng Huang
}

\author[$\ddag$]{
Jincheng Yu
}

\affil[$\ddag$]{Department of Computer Science and Engineering\\ HKUST}
\affil[$\dag$]{Huawei Noah's Ark Lab}
\affil[ ]{\{jzhangey, xhuangck, jyubh\}\textit{@}connect.ust.hk}

\date{}
\maketitle



\begin{abstract}%
We consider optimizing two-layer neural networks in the mean-field regime where the learning dynamics of network weights can be approximated by the evolution in the space of probability measures over the weight parameters associated with the  neurons. The mean-field regime is a theoretically attractive alternative to the NTK (lazy training) regime which is only restricted locally in the so-called neural tangent kernel space around specialized initializations. Several prior works (\cite{chizat2018global, mei2018mean}) establish the asymptotic global optimality of the mean-field regime, but it is still challenging to obtain a quantitative convergence rate due to the complicated unbounded nonlinearity of the training dynamics.  This work establishes the first linear convergence result for vanilla two-layer neural networks trained by continuous-time noisy gradient descent in the mean-field regime. Our result relies on a novel time-depdendent estimate of the logarithmic Sobolev constants for a family of measures determined by the evolving distribution of hidden neurons. 
\end{abstract}

\section{Introduction}
Gradient-based optimization is a fundamental tool in machine learning and has witnessed great empirical success for training neural networks, despite the highly non-convexity landscape of the objective. However, theoretical understanding of nonconvex optimization in neural networks is quite limited. Until recently, there has been much work that explains the success of gradient-based optimization in overparametrized neural networks, that is neural networks with massive hidden units. Under the overparametrization condition, the learning problem can be translated into minimizing a convex functional and hence circumventing the difficulties of analyzing non-convex objectives. 

It's worth mentioning that there has been much broader interest in analyzing the convergence of machine learning algorithms by formulating it as the problem of minimizing some (usually convex) functional of a measure, such as variational inference (\cite{liu2016stein, liu2017stein, chewi2020svgd}), generative adversatial networks (\cite{johnson2019framework, nitanda2020functional}) and learning infinite-width neural networks (\cite{chizat2018global, mei2018mean, nguyen2020rigorous, fang2021modeling}).  The key idea is by approximating the learning dynamics of model parameters by the optimization on the space of probability of measures over the model parameters under the overparametrization condition. Such functional optimization procedure corresponds to the gradient flow over the space of probability measure endowed with some probability metrics such as the Wasserstein distance (\cite{chizat2018global, mei2018mean}). The functional optimization framework enables us to study the optimization and generalization of learning algorithms with a different set of tools, such as non-linear PDEs for optimial transport (\cite{ambrosio2008gradient, villani2009optimal, santambrogio2015optimal}).

\subsection{Related Works}\label{related}
The study of overparametrized neural networks mainly falls into two categories: the NTK regime or the mean-field regime. 

\textbf{NTK Analysis of Neural Networks.} It has been already known in the literature that  deep neural network in the infinite-width limit is equivalent to a Gaussian process since decades ago (\cite{neal1996priors, williams1997computing, winther2001computing, neal2012bayesian, lee2018deep}) and thus they can be connected to kernel methods. Later, \cite{jacot2018neural} shows that the evolution of the trainable parameters in continuous width deep neural nets during training process can be captured by a kernel, which is referred to Neural Tangent Kernel (NTK). Recent work has shown that with a specialized scaling and random initialization, the parameters of continuous width two-layer neural network is restricted in an infinitesimal region around the initialization and can be regarded  a linear model with infinite dimensional feature (\cite{du2018gradient, li2018learning, arora2019fine, du2019gradient, li2020learning}). Since the system becomes linear, the dynamics of GD within this region can be tracked via properties of the associated NTK and the convergence to the global optima with a linear conergence rate can be proved. Later, \cite{allen2018learning,  zou2018stochastic, allen2019convergence} extend the NTK analysis to the global convergence of GD for multi-layer neural nets. Despite the huge theoretic progress of NTK regime in analyzing both shallow and deep neural networks, such regime deviates far from practical fully trained neural networks. For NTK approximation, the training is restricted to a small region around the initialization and therefore it suffers from lazy training and cannot explain why neural networks learn discriminative features in practice. An alternative and more promising approach is to consider the mean-field modeling of neural networks. 

\textbf{Mean-Field Analysis of Neural Networks.} Another strand of work tries to study the infinite-width neural network from a mean field view (\cite{chizat2018global, mei2018mean,  sirignano2019mean, nguyen2020rigorous, fang2021modeling}). The key idea is to characterize the learning dynamics of SGD (\cite{chizat2018global}) or noisy SGD (\cite{mei2018mean}) as the gradient flow over the space of probability distribution of NN parameters. More specifically, it has been proved in \cite{chizat2018global} that SGD enjoys asymptotic global optimality under homogeneous activations in the mean-field regime. While independently in \cite{mei2018mean}, the asymptotic global optimality relies on the mixing effect of noise, hence the analysis mainly focuses on the noisy SGD. In this paper, we analyze the noisy SGD and hence adopt the framework in \cite{mei2018mean} that the learning dynamics of noisy SGD can be characterized by a non-linear PDE of Mckean-Vlason type, which corresponds to the Wasserstein gradient flow over the space of distributions of NN parameters. When the time goes to infinity, the noisy SGD convergences asymptotically  to the unique global optima of the convex objective functional for two-layer NN with infinite width.

Despite lots of advances in analyzing the dynamics and global convergence of neural networks trained by noisy SGD in the mean-field regime (\cite{mei2018mean, chizat2018global}), these results are qualitative without known convergence rates. Hence, an important research question arises: 
\begin{center}
\emph{Can we obtain a quantitative convergence rate for overparametrized \\ neural networks trained by noisy SGD in the mean-field regime?}
\end{center}
This question remains open even for the simple vanilla two-layer neural networks. In this paper, we address this question and prove the first linear convergence result for standard two-layer neural networks trained by continuous-time noisy gradient descent in the mean-field regime.

Prior to the submission of this work, we are aware that several recent independent works (\cite{chizat2022mean, nitanda2022convex}) have been on Arxiv that also tried to prove the linear convergence result for neural networks trained by noisy SGD in the mean-field regime, but as pointed out by the authors (\cite{chizat2022mean}), their assumptions of boundedness and smoothness in both works cannot be applied to the vanilla two-layer neural networks (we will discuss details in later sections).  Our work differs from them in both assumptions and proof techniques. Specifically,
\begin{itemize}
\item In \cite{chizat2022mean, nitanda2022convex}, it is assumed that each particle is bounded and smooth with respect
both weight and position of each neuron. This assumption is unrealistic even for vanilla two-layer neural networks unless we fix the parameter of the second layer and assume the activation is bounded. In such case, it is not a real two-layer neural network. 

\item As we will see later in this work, the unboundedness and non-smoothness of each particle with respect to its parameter is the main technical hurdle to prove the linear convergence result. Unlike \cite{chizat2022mean, nitanda2022convex} which directly make restrictive assumptions on the function of neural networks, we address this technical hurdle by imposing regularizers with particular growth rate and hence it can control the curvature of the whole objective. The regularizer is similar to weight decay and can be easily implemented.

\item Our technique could yield a more refined time-dependent estimate of the logarithmic sobolev constant and if it is uniformly upper bounded for all $t\geq 0$, then such an upper bound can be the linear convergence rate. While in \cite{chizat2022mean, nitanda2022convex}, it is assumed that the logarithmic Sobolev inequality uniformly holds for some set of measures in $\mathcal{P}_2(\mathbb{R}^d)$ and the estimate of Sobolev constant is directly obtained from assumptions which is independent of the time. Such rough constant may not reveal the real convergence rate of the algorithm. 
\end{itemize}
There are other existing works that also show the linear convergence of two-layer neural networks in the mean-field regime. More specifically, \cite{chen2022feature} show that under the non-degeneracy condition on certain Gram matrix, gradient descent (GD) can linear converge to $0$ training loss efficiently in the mean-field scaling. Our approach is not directly comparable to \cite{chen2022feature} since the analysis is an extension of NTK-theory to the feature learning regime  that is different from the continuous noisy SGD in the mean-field regime studied in our paper. A more related work is \cite{hu2019mean} which obtain the linear convergence of noisy SGD for two-layer NN in the mean-field regime, but the convergence result requires the sufficiently strong regularization and does not exploit the convexity of the functional of neural networks. In some sense, the convergence in \cite{hu2019mean} is only because of sufficiently strong regularizations. While in our analysis, the regularization could be arbitrarily small.

\subsection{Contribution}
In this paper, we prove that under some mild regularity conditions on the regularizer, the solution of the noisy SGD on a two-layer neural network converges linearly to the global optima of the regularized objective in the mean-field regime. More specifically, 
\begin{itemize}
\item We present an analysis to show that logarithmic Sobolev inequality can be directly applied to mean-filed analysis of two-layer neural networks. If a uniform upper bound on the logarithmic Sobolev constants can be obtained for a family of measures determined by the evolving distribution of hidden neurons along the trajectory, then we obtain a linear convergence rate in the mean-field regime. 

\item We develop new techniques to estimate the upper bound of logarithmic Sobolev constants for measures determined by the evolving distribution of parameters of two-layer neural networks. We show that for some carefully chosen regularizer, an upper bound on the logarithmic Sobolev constants can be obtained and hence 
we can achieve the linear convergence result for the vanilla two-layer neural networks for the first time. This solves the unboundedness issue raised by \cite{chizat2022mean}, where the authors claim that there should be incompatibility between noisy gradient descent and the standard architecture of two-layer neural networks. We instead show this should be compatible if we properly chose the regularizer. 
\end{itemize}
Our theory also has practical implications: different regularizers could lead to different estimate of the log-Sobolev constants, hence influencing the convergence rates to the global optima. Designing good regularizers for effective training neural networks with theoretic guarantees is a promising direction for future research. 

\subsection{Notations}
Let $\|\cdot\|_2$ denote the Euclidean norm of a vector and $\|\cdot\|_{op}$ denote the operator norm of a matrix. We use $\nabla$, $\nabla\cdot$, $\Delta: =\nabla\cdot\nabla$ and $\nabla^2$ to denote the gradient, divergence, Laplacian, and Hessian operator respectively.  $\langle \cdot, \cdot\rangle$ denotes the inner product between vectors or functions. For $\mu, \nu\in\mathbb{P}_2(\mathbb{R}^d)$, the $2$-Wasserstein distance is defined as
\begin{eqnarray}
&&W_2(\mu, \nu)^2 := \min_{\gamma\in\Pi(\mu, \nu)}\int \|x-y\|_2^2\gamma(dx, dy)~
\end{eqnarray}
where $\Pi(\mu, \nu)$ denotes the coupling between $\mu$ and $\nu$, which means $\Pi(\mu, \nu)$ is a measure on $\mathbb{R}^d\times \mathbb{R}^d$ with marginals $\mu$ and $\nu$ respectively.

\section{Problem Setup}\label{section3}
For an input feature vector $x\in\mathbb{R}^d$, consider a two-layer neural network with $N$ hidden neurons
\begin{eqnarray}
\hat{f}_N(x, \Theta) = \frac{1}{N}\sum_{i=1}^{N}\tilde{h}(\theta_i, x) =  \frac{1}{N}\sum_{i=1}^{N}u_ih(w_i, x)~.
\end{eqnarray}
Here, $N$ is the number of hidden neurons; $\Theta = [\theta_1,\ldots, \theta_N]\in \mathbb{R}^{(d+1)\times N}$ are parameters of the neural network; each $\theta_i= (u_i, w_i)\in\mathbb{R}^{d+1}$ denotes a weighted neuron, with position $w_i\in\mathbb{R}^{d}$ and weight $u_i\in \mathbb{R}$; $h(\cdot, \cdot)$ is the (nonlinear) activation function. 

In the mean-field regime, each weighted neuron $\theta_i$ can be regarded as a particle that is drawn i.i.d. from a distribution $\rho(\theta)$.  When the number of hidden neurons tends to infinity, we obtain the mean-field limit of the two-layer neural network
\begin{eqnarray}
f(x, \rho) = \int_{\mathbb{R}^{d+1}} \tilde{h}(\theta, x)\rho(\theta)d\theta~.
\end{eqnarray}

In this paper we consider the classical setting of supervised learning whereby we are given a feature-label pair $(x, y) \in \mathbb{R}^d
\times \mathbb{R}$ drawn from an unknown distribution. If we choose the loss function $\phi: \mathbb{R}
\times \mathbb{R}\to  \mathbb{R}$, we wish to minimize the expected loss over the data distribution. In order to control the model capacity, we also add some regularizers to the parameters of the neural network. Then the objective of the two-layer neural network in the mean-field limit becomes:
\begin{eqnarray} \label{obj}
Q(\rho) = \mathbb{E}_{x, y}\phi\left(\int_{\mathbb{R}^{d+1}} \tilde{h}(\theta, x)\rho(\theta)d\theta, y\right) + \int_{\mathbb{R}^{d+1}} r(\theta)\rho(\theta)d\theta + \lambda \int_{\mathbb{R}^{d+1}} \rho(\theta)\log \rho(\theta)d\theta 
\end{eqnarray}
where $\tilde{h}(\theta, x) = uh(w, x)$ with $u\in\mathbb{R}$ and $\theta = [u, w]\in \mathbb{R}^{d+1}$; $\rho(\theta) $ is a probability distribution of neurons; $r(\theta)$ is the regularizer that penalizes the weight norm of neurons. 
\begin{remark}
The above formulation of infinite-width two-layer neural networks can be easily implemented by i.i.d sampling the weight $\theta$ of neurons over the distribution $\rho(\theta)$. For the second term, it corresponds to the weight decay when $r(\theta)$ is the square of the Euclidean norm of $\theta$.  For the last entropy-regularized term,  it can be implemented by adding Gaussian noise, which corresponds to the noise term in the training of the neural network weights by noisy SGD.
\end{remark}
\begin{remark}
In \cite{chizat2022mean} and \cite{nitanda2022convex}, it is assumed that $\tilde{h}(\theta, x)$ is bounded and smooth with respect to $\theta$ uniformly for any $x$. This assumption is restrictive and cannot be applied in two-layer neural networks, since we could not control the boundedness of $\tilde{h}(\theta, x)= uh(w, x)$ with respect to $u$. Even when we add another bounded and smooth activation $g$ in the output, and let $\tilde{h}(\theta, x)= g(uh(w, x))$, the smoothness of $\tilde{h}(\theta, x)= g(uh(w, x))$ with respect to $\theta$ still cannot hold. As shown in our proof, the spectral norm of the Hessian $\nabla_\theta^2 \tilde{h}(\theta, x)$ will tend to infinity as $|u|\to\infty$. 
\end{remark}

\subsection{The Decreasing Path Determined by the Gradient Flow}
 In order to minize the regularized objective (\ref{obj}), we consider the following particle optimization for neurons
\begin{eqnarray}
d\theta_t = -\nabla_{\theta}g(\theta_t,  \rho_t)dt~.
\end{eqnarray}
It's straightforward to get the following as the standard result of the Fokker-Planck equation:
\begin{eqnarray}\label{eqn1.3}
\frac{\partial \rho_t(\theta)}{\partial t} = \nabla\cdot[\rho_t\nabla g(\theta, \rho_t)].
\end{eqnarray}

Then, we obtain the following decreasing lemma, which is standard and well-known in the existing literature by simply integral-by-parts. 
\begin{lemma}[The Decreasing Lemma] \label{lemma_dec}
If we take
\begin{eqnarray}
&&g(\theta, \rho) =U(\theta, \rho) + \lambda \log \rho(\theta)\quad \text{and} \quad  U(\theta, \rho) = \mathbb{E}_{x, y}\phi_1^\prime(f(\rho, x), y)\tilde{h}(\theta, x) + r(\theta)
\end{eqnarray}
where $f(\rho, x) = \int_{\mathbb{R}^{d+1}} \tilde{h}(\theta, x)\rho(\theta)d\theta$ and $\phi_1^\prime(\cdot, \cdot)$ is the derivative of $\phi$ with respect to its first argument.
We can get
\begin{eqnarray}
&& \frac{dQ(\rho_t)}{dt}= -\int_{\mathbb{R}^{d+1}} \|\nabla_{\theta} g(\theta, \rho_t)\|_2^2\rho_t(\theta)d\theta~.
\end{eqnarray}
\begin{proof}
By definition and integral by parts, we obtain
\begin{eqnarray}
&&  \frac{dQ(\rho_t)}{dt} = \left \langle \frac{\delta Q(\rho_t) }{\delta \rho_t},  \frac{\partial \rho_t(\theta)}{\partial t} \right\rangle =  \left \langle g(\theta, \rho_t),    \nabla\cdot[\rho_t\nabla g(\theta, \rho_t)]\right\rangle = - \left \langle \|\nabla_{\theta} g(\theta, \rho_t)\|_2^2,    \rho_t \right\rangle ~.\nonumber
\end{eqnarray}
\end{proof}
\end{lemma}
\begin{remark}
Notice that in Lemma $\ref{lemma_dec}$, we have picked $$g(\theta, \rho)= \mathbb{E}_{x, y}\phi_1^\prime(f(\rho, x), y)\tilde{h}(\theta, x) + r(\theta) + \lambda \log \rho(\theta)~,$$ which is in fact the linear functional derivative of the objective $Q(\rho)$. Then the particle optimization for neurons becomes
\begin{eqnarray} \label{eqn_log} 
d\theta_t = -\nabla_{\theta}U(\theta_t,  \rho_t)dt -\lambda\nabla \log \rho_t(\theta) dt~
\end{eqnarray}
and the corresponding nonlinear Fokker-Planck equation becomes
\begin{eqnarray} \label{nonlinear}
\frac{\partial \rho_t(\theta)}{\partial t} = \nabla\cdot[\rho_t\nabla U(\theta, \rho_t)] + \lambda \Delta \rho_t(\theta).
\end{eqnarray}
Notice that  the particle update (\ref{eqn_log}) is hard to implement due to the term $-\lambda\nabla \log \rho_t(\theta) dt$, which can be replaced by adding diffusions, i.e.,
\begin{eqnarray} 
d\theta_t = -\nabla_{\theta}U(\theta_t,  \rho_t)dt +\sqrt{2\lambda}dB_t~
\end{eqnarray}
where $dB_t$ is the standard Brownian motion. 
\end{remark}
\begin{remark}
The decreasing lemma shows that the risk will be non-increasing along the gradient flow dynamics. It is quite standard in the literature of Wasserstein gradient flows (e.g., see Section $4.4.E$ of \cite{ambrosio2007gradient}) where one needs the existence of the first-order variations and the convexity along the Wasserstein geodesics for the objective functional , which can be easily verified in our case. 
\end{remark}
The well-posedness of the nonlinear Fokker-Planck equation for standard two-layer neural networks has been proved in \cite{mei2019mean}, where the potentials are not globally bounded and Lipschitz anymore.
\begin{proposition}[Generalization of the Proposition $51$ in \cite{mei2019mean}]\label{well}
Let the initialization $\rho_0\in\mathbb{P}_2(\mathbb{R}^{d+1})$ such that $Q(p_0)<\infty$. Then, the solution $(\rho_t)_{t\geq 0}\in\mathbb{P}_2(\mathbb{R}^{d+1})$ of (\ref{nonlinear}) exists and is unique under some mild smoothness and growth conditions on the regularizer $r(\theta)$. 
\end{proposition}

\subsection{Assumptions}
Before we introduce our main result, here we provide several assumptions on the loss function $\phi(\cdot, \cdot)$ and the activation function $h(\cdot, \cdot)$ that are critical in our subsequent analysis. 
\begin{assumption}
The loss function $\phi(\cdot, \cdot)$ satisfies the following conditions:
\begin{itemize}
\item (Convexity.) $\phi(\cdot, \cdot)$ is convex on its first argument.
\item (Lower Boundedness.) $\phi(\cdot, \cdot)$ is bounded below, i.e., $\exists B_l>0$ ~such that ~$\forall y'\in\mathbb{R}$ and  $y\in \mathbb{R}$, $\phi(y', y)\geq B_l$~.
\item (Lipschitz Gradient.)  $\phi(\cdot, \cdot)$ is $L_1$-bounded and $L_2$-Lipschitz continuous gradient on its first argument, i.e., $\exists L_1, L_2>0$ such that $\forall y_1, y_2\in\mathbb{R}$ and $y\in\mathcal{Y}$, $|\phi_1'(y_1, y)|\leq L_1$ and $$|\phi_1'(y_1, y)-\phi_1'(y_2, y)|\leq L_2|y_1-y_2|~.$$
\end{itemize}
\end{assumption}
Note that such assumptions on the loss function are commonly used in existing literatures (\cite{du2019gradient}). We also need some assumptions on the activation function.
\begin{assumption}
The feature activation function satisfies the following conditions:
\begin{itemize}
\item (Differentiability.) $\forall x\in\mathcal{X}$, $h(\cdot , x)$ is second-order differentiable. 
\item (Growth Condition.) $\forall x\in\mathcal{X}$ and $w\in\mathbb{R}^d$, $|h(w, x)|\leq C_1\|w\|+C_2~,$ $\|\nabla_1h(w, x)\|\leq C_3~,$ and $\|\nabla_1^2 h(w, x)\|_{op}\leq C_4$ for some constants $C_1, C_2, C_3, C_4>0$~.
\end{itemize}
\end{assumption}
The assumptions on the activation function also incorporate many commonly used activations, such as sigmoid and smoothed ReLU.

\section{Global Optimality with Linear Convergence Rates}
We need to introduce some functional inequalities that is essential in the proofs of theorems.
\subsection{Poincare and Logarithmic Sobolev Inequality}
Consider a probability space $(\mathcal{X}, \mathcal{F}, \mu_t)$ where $\mu_t$ is some measure at time $t\geq 0$ and let $\mathcal{L}_t$ be a self-adjoint operator on $L^2(\mu_t)$ with domain $\mathbb{D}_2(\mathcal{L}_t)$ which is a dense subset of $L^2(\mu_t)$ and the corresponding Dirichlet form associated to $\mathcal{L}_t$ is 
\begin{eqnarray}
&& \mathcal{E}_t(f) = -\langle \mathcal{L}_tf, f\rangle_{\mu_t}, \quad f\in \mathbb{D}_2(\mathcal{L}_t)~.
\end{eqnarray}

In our setting, we consider the symmetric diffusion in the Euclidean space. More specifically, we let $\mathcal{X}=\mathbb{R}^{d+1}$ and let $\mathcal{L}_t = \nabla\cdot\nabla - \nabla U_t\cdot\nabla$ for some $U_t\in W^{1, 2}_{loc}(\mathbb{R}^{d+1})$ such that $Z = \int_{\mathbb{R}^{d+1}}e^{U_t(\theta)}d\theta<\infty$ and $d\mu_t(\theta) = Z^{-1}e^{-U_t(\theta)}d\theta$. In this case, the associated Dirichlet form is given by the symmetric property as follows
\begin{eqnarray}
&& \mathcal{E}_t(f)  = \int_{\mathbb{R}^{d+1}}\left\|\nabla f(\theta)\right\|_2^2d\mu_t(\theta)~.
\end{eqnarray}

Throughout the paper, if it is not mentioned explicitly, all test functions in the functional inequalities should be in $\mathbb{D}_2(\mathcal{L}_t)$. It's well-known that the spectral gap exists for $\mathcal{L}_t$, if and only if the following Poincare inequality holds
\begin{eqnarray}
&& Var_{\mu_t}(f) := \int f^2 d\mu_t - \left(\int f d\mu_t\right)^2 \leq C_{P_t}  \mathcal{E}_t(f) 
\end{eqnarray}
where $C_{P_t}^{-1}:= \lambda_t$ is the spectral gap, which is defined as
\begin{eqnarray}
&&\lambda_t := \inf\left\{  \frac{\mathcal{E}_t(g) }{\|g\|_{L^2(\mu_t)}^2}: g\in C^2, g\neq 0, \int g =0 \right\}~.
\end{eqnarray}

A defective logarithmic Sobolev inequality (DLSI) is satisfied if 
\begin{eqnarray}
&& Ent_{\mu_t}\left(f^2\right) :=  \int f^2 \log f^2 d\mu_t - \int f^2 d\mu_t  \log\left(\int f^2 d\mu_t\right) \leq \nu_{\mu_t} \mathcal{E}_t(f)  + D_{LS_t} \int f^2 d\mu_t. \nonumber
\end{eqnarray}

The logarithmic sobolev inequality (LSI) holds when $D_{LS_t} = 0$ and the smallest $\nu_{\mu_t}$ for which this inequality holds is said to be the log-Sobolev constant for $\mu_t$. It's worth mentioning that from the Rothaus’s lemma (See Theorem $5.1.4$ in \cite{bakry2014analysis}), if the spectral gap exists for $\mathcal{L}_t$, a defective LSI, can be transformed into a LSI (see \cite{barthe2008mass} and references therein).

\subsection{Global Linear Convergence via Uniform Log-Sobolev Inequalities}
Now, we are ready to present the following proposition, which obtains a linear convergence of $Q(\rho_t)$ to its global optima $Q(\rho_*)$ under the condition that the logarithmic sobolev inequality is satisfied uniformly for a family of Gibbs measures satisfying $\tilde{\rho}_t \propto \exp(-\lambda^{-1}U(\theta, \rho_t))$.

\begin{proposition}
The functional $Q(\rho)$ has a unique global optimal solution $\rho_*$.  Furthermore, when the logarithmic sobolev inequality is satisfied uniformly with a coefficient $\nu>0$ for all measures $\tilde{\rho}_t \propto \exp(-\lambda^{-1}U(\theta, \rho_t))$ with $t\geq 0$, then $Q(\rho_t)$ is monotonically non-increasing with $\rho_t$ converges weakly to $\rho_*$. Furthermore, we have for $t\geq 0$ and any $\rho_0$ with $Q(\rho_0)<\infty$,
\begin{eqnarray}
Q(\rho_t)-Q(\rho_*)\leq [Q(\rho_0)-Q(\rho_*)]*\exp(-2\lambda\nu^{-1} t)
\end{eqnarray}
where $\nu >0$ is a problem dependent constant.

\begin{proof}
Since the functional $Q(\rho)$ is convex with respect to $\rho$, it follows from standard arguments in the literature that there is a unique global optimal solution $\rho_*$ to $Q(\rho)$ which $\rho_t$ weakly converges to (\cite{mei2018mean}).

By the decreasing lemma, we have
\begin{eqnarray}
\frac{dQ(\rho_t)}{dt}= -\int_{\mathbb{R}^{d+1}} \|\nabla_{\theta} g(\theta, \rho_t)\|_2^2\rho_t(\theta)d\theta = -\lambda^2\int_{\mathbb{R}^{d+1}} \|\nabla_{\theta}\log\exp(\lambda^{-1} g(\theta, \rho_t))\|_2^2\rho_t(\theta)d\theta~. 
\end{eqnarray}
If the logarithmic sobolev inequality is satisfied uniformly with a problem dependent coefficient $\nu>0$ for all measures $\tilde{\rho}_t \propto \exp(-\lambda^{-1}U(\theta, \rho_t)) = \frac{\rho_t(\theta)\exp(-\lambda^{-1}g(\rho_t, \theta) )}{\mathbb{E}_{\theta\sim\rho_t} \exp(-\lambda^{-1}g(\rho_t, \theta) )}$.
Therefore, we obtain for all $t\geq 0$, $ Ent_{\tilde{\rho}_t}\left(f^2\right)\leq \nu \mathcal{E}_t(f)$ ~.
If we set $f(\theta) = \sqrt{\rho_t(\theta)/\tilde{\rho}_t(\theta)} \in L^2(\tilde{\rho}_t)$, we obtain
\begin{eqnarray}
2\nu^{-1} KL(\rho_t\|\tilde{\rho}_t) \leq \mathbb{E}_{\theta\sim \rho_t}\left\| \nabla\log\frac{\rho_t(\theta)}{\tilde{\rho}_t(\theta)}\right\|_2^2~.
\end{eqnarray}
Therefore, we obtain
\begin{eqnarray}
&&  \frac{dQ(\rho_t)}{dt}\leq -2\lambda^2\nu^{-1}\mathbb{E}_{\theta\sim \rho_t}\log\frac{\mathbb{E}_{\theta\sim\rho_t} \exp(-\lambda^{-1}g(\rho_t, \theta) ) }{  \exp(-\lambda^{-1}g(\rho_t, \theta)) }~.
\end{eqnarray}
Since $\phi(\cdot, \cdot)$ is convex on its first argument, we obtain that $Q(\rho)$ is a strongly convex functional with respect to $\rho$, and it is straightforward to obtain, 
\begin{eqnarray}
&& Q(\rho_t) - Q(\rho_*) \leq \int_{\mathbb{R}^{d+1}} g_t(\theta, \rho_t)(\rho_t(\theta)-\rho_*(\theta))d\theta -\lambda KL(\rho_*\|\rho_t) \nonumber\\
&&  \leq \lambda\log\mathbb{E}_{\theta\sim\rho_t}\exp(-\lambda^{-1}g_t(\theta, \rho_t)) + \mathbb{E}_{\theta\sim\rho_t}g_t(\theta, \rho_t)~.
\end{eqnarray}
It follows that
\begin{eqnarray}
&& \frac{d[Q(\rho_t) - Q(\rho_*)]}{dt} \leq -2\lambda\nu^{-1} [Q(\rho_t) - Q(\rho_*)]~
\end{eqnarray}
which implies the linear convergence of $Q(\rho_t)$ to $Q(\rho_*)$~.
\end{proof}
\end{proposition}

\begin{remark}
We shall note that there are two Arxiv paper (\cite{chizat2022mean, nitanda2022convex}) that already obtain the above results prior to the submission of this manuscipt. Our work is done independently to them and the difference is that we assume the uniform LSI holds along the trajectory rather than for any probability measure in $\mathcal{P}_2(\mathbb{R}^d)$ as in (\cite{chizat2022mean, nitanda2022convex}). Although this weaker assumption is straightforward from a technical perspective, we stress it because it can lead to a more refined time-dependent estimate of LSI constants in Theorem \ref{main3}.\\
From a higher perspective, the difficulty to obtain the convergence rate arises from the nonlinearity in the drift term $ -\nabla_{\theta}U(\theta_t,  \rho_t)$ of the continuous stochastic process corresponds to the Mckean-Vlason PDE, in contrast to the classical linear Fokker-Planck equation that describes the evolution dynamics of Langevin algorithm where the drift term remains unchanged during the evolution and if we put some isoperimetry condition on the stationary distribution determined by the drift term, such as it satisfies the Logarithmic Sobolev Inequality (LSI), then the linear convergence  to the stationary process can be obtained and the convergence rate can be controlled by the LSI constant of the stationary distribution (\cite{ma2019sampling}). To overcome such difficulty, one can simply adapt the proofs in (\cite{ma2019sampling}) and substitute the stationary process $p^* \propto \exp(-\lambda^{-1}U(\theta))$ by the varying Gibbs distribution $\tilde{\rho}_t \propto \exp(-\lambda^{-1}U(\theta, \rho_t))$. Then the convergence rate can be controlled by the uniform Log-Sobolev constant of $\tilde{\rho}_t$ along the optimization trajectory.
\end{remark}
\begin{remark}
In a broader context, the idea of assuming uniform LSI to obtain the linear convergence of McKean-Vlasov process already exists in the literature, for example, in \cite{yang2020variational} and \cite{guillin2019uniform}. The fundamental difference between these works and our contribution is that these works directly assume the uniform LSI holds while we prove that the uniform LSI holds in the context of vanilla two-layer neural networks.
\end{remark}

\section{Estimation of the Uniform Log-Sobolev Constant}
In the previous section, we have obtained the linear convergence rate for noisy SGD in training two-layer neural networks in the mean-field regime. But the whole story is far from complete, since this linear convergence result relies on the condition that $\tilde{\rho}_t \propto \exp(-\lambda^{-1}U(\theta, \rho_t))$ satisfies the LSI with coefficients $\nu_t$ and there exist a uniform upper bound $\nu>0$ on $\nu_t$  for all $t\geq 0$. Hence, in the next, we need to estimate the upper bound of the log-Sobolev constants for $\tilde{\rho}_t \propto \exp(-\lambda^{-1}U(\theta, \rho_t))$.

There are many literatures on establishing the logarithmic sobolev constant for Gibbs distributions of the form $\mu_t\propto e^{-U_t}$. The most simple and fundamental one is the Bakery-Emery Criteria (\cite{bakry1985diffusions}): if $U_t$ is $C^2$ and strongly convex, i.e., $\nabla^2 U_t \succeq K I$ for some $K>0$, then $\mu_t$ satisfies the LSI with $\nu_{\mu_t} = 1/K$. However, in many cases, $U_t$ is nonconvex. 

In nonconvex cases, the standard tool to deduce the the LSI is the Holley-Stroock perturbation principle (\cite{holley1986logarithmic}). The idea is that if we can find a strongly convex function $\hat{U}_t$ that is a bounded perturbation of $U_t$, then $U_t$ satisfies the LSI with a worse logarithmic Sobolev constant than $\hat{U}_t$. From an intuitive understanding, in this case, $U_t$ should still be dominated by a strongly convex function, plus some minor bounded nonconvex function as perturbations. This method is adopted in some machine learning literatures of analyzing the Langevin sampling with nonconvex potentials, for example, see \cite{ma2019sampling}~, where $U_t$ is strongly convex outside a ball with fixed radius. In this case,  we can construct a globally strongly convex function $\hat{U}_t$ that is the bounded perturbation of $U_t$, hence deducing the LSI. 

There are some other extensions to obtain the LSI beyond Bakery-Emery criteria. For example, if there is a constant lower bound on the Hessian, i.e., $\nabla^2 U_t\succeq -KI$ for some $K>0$, and some Lyapunov function criteria is satisfied, then we can also obtain the LSI for $\mu_t$. This method also appears in analyzing the global convergence of SGLD, for example, see \cite{raginsky2017non}.

There are some recent works that establish the logarithmic sobolev inequality when the Bakery-Emery Curvature is not lower bounded, which is considered in \cite{barthe2008mass} and later \cite{cattiaux2010note} extends the results of \cite{barthe2008mass} in Euclidean space to the general Riemann Manifold. To our best knowledge, this approach has not been considered in any machine learning literature. 

However, it is generally hard to estimate the log-Sobolev constants in the absence of convexity, which is the case when $U_t$ is a nonconvex potential for optimizing a two-layer infinite-width neural network. As will shown in the later analysis, the key technical difficulty is that the lower bound of the Hessian of the two-layer infinite-width neural network grows linearly in terms of the norm of the neuron. How can we overcome such difficulty? Notice that we have adopted a regularizer $r(\theta)$: if the Hessian $\nabla^2 r(\theta)$ grows sufficiently fast as a function of $\|\theta\|_2$, then it may help us control the Hessian of $U_t$. Observe that we cannot adopt the original weight decay $r(\theta) = \frac{\beta}{2}\|\theta\|_2^2$, since the Hessian $\nabla^2 r(\theta)=\beta I$, which is a constant and cannot control the Hessian of the whole $U_t$. This motivates us to consider higher order regularizers beyond weight decay.

In the next two subsections, we consider two cases of $r(\theta)$ where we can bound the log-Sobolev constants uniformly for all $\mu_t\propto e^{-U_t}$.

\subsection{Estimate of the Logarithmic Sobolev Constant via Holley-Stroock Perturbation Principle}
First, we consider a special case $r(\theta) = \beta\| \theta\|_2^4$, which can be regarded as a variant of weight decay by enhancing more decay on weights with larger Euclidean norm. In such case, $U(\theta, \rho_t)$ will be strongly convex outside a ball with some radius $R_{m, \beta}>0$ and locally smooth on any compact set in $\mathbb{R}^d$. Using similar construction as \cite{ma2019sampling} and then adopting the Holley-Stroock perturbation principle, we can bound the LSI constant uniformly for all $\mu_t\propto e^{-U_t}$. As will seen in the proof, our analysis can be directly extended to any $r(\theta)$ in $C^2$ that grows not slower than $\|\theta\|_2^4$. 
\begin{theorem} [Proved in Appendix \ref{A}] \label{main2}
When $r(\theta)= \beta \| \theta\|_2^4$, we have the following uniform upper bound on the Logarithmic Sobolev constant $\nu_{\rho_t}$, 
\begin{eqnarray}
&&\sup_{t\geq 0}\nu_{\tilde{\rho}_t} \leq \frac{2}{m\lambda^{-1}} e^{16 \lambda^{-1}L_{m, \beta} R_{m, \beta}^2 }~.\nonumber
\end{eqnarray}
where $m>0$  can be picked arbitrarily and $R_{m, \beta},  L_{m, \beta}>0$ are all problem dependent constants (detailed in appendix).
\end{theorem}

\subsection{Time-dependent Estimate of the Logarithmic Sobolev Constant}
Notice that when we adopt the Holley-Stroock perturbation principle, the obtained LSI could be too loose since there is a term which grows exponentially with the perturbation. This estimate of Log Sobolev does not depend on the trajectory dynamics, which is also a weakness in existing works (\cite{chizat2022mean, nitanda2022convex}). Furthermore, the condition that regularizer $r(\theta)$  grows at least $\|\theta\|_2^4$ fast could be too strong.  This motivates us to seek alternative and more general approaches to provide a more refined estimate on the LSI constants: we will show, under some Lyapunov conditions together with a uniform spectral gap, the LSI holds uniformly for all $\mu_t\propto e^{-U_t}$ for any $r(\theta)$ in $C^2$ that grows at least as fast as $\|\theta\|_2^r$ for any $r\geq 3$. Notice that in this case the Bakery-Emery Curvature for $\mu_t$ is not lower bounded, hence we adopt the techniques in \cite{barthe2008mass, cattiaux2010note}.

We need to put some regularity conditions on the regularizer $r(\theta)$, that is critical in establishing the Lyapnov conditions for the Logarithmic Sobolev ineuqality.
\begin{assumption} \label{assumption_reg}
The regularizer $r(\theta)$ should satisfy
\begin{itemize}
\item For some $m> 0$, $b\geq 0$ and $p\geq 1$,  
\begin{eqnarray}
&&\langle \theta,\nabla r(\theta) \rangle \geq m\|\theta\|_2^{2+p} -b, \quad \forall \theta\in\mathbb{R}^{d+1}~.
\end{eqnarray}
\item For some $D_1, D_2, D_3, D_4, D_5, D_6, D_7, D_8>0$, and $k\geq 3$
\begin{eqnarray}
&& (D_1\|\theta\|_2 +D_2)I  \preceq \nabla^2 r(\theta) \preceq (D_3\|\theta\|_2^p +D_4)I,
\end{eqnarray}
and
\begin{eqnarray}
&& \|\nabla r(\theta)\|_2 \geq D_5\|\theta\|_2^2+D_6,
\end{eqnarray}
and 
\begin{eqnarray}
&& |r(\theta)| \leq D_7\|\theta\|_2^k +D_8~.
\end{eqnarray}
\end{itemize}
\end{assumption}
\begin{remark}
The above assumptions naturally apply to many natural regularizers such as $r(\theta) =\frac{\beta_1}{2}\|\theta\|_2^2+ \frac{\beta_2}{3}\|\theta\|_2^3~,$ which also appears in the analysis of \cite{allen2020feature}.
\end{remark}

As will be shown in the proof, we will first establish a time-dependent estimate of the Logarithmic Sobolev constants $\nu_{\rho_t}$ for a family of Gibbs distributions with potential $U(\theta; \rho_t)$. Then, for $t\geq 0$, we have the following uniform upper bound for $\nu_{\rho_t}$.
\begin{theorem}[Proved in Appendix \ref{B}] \label{main3}
We have the following uniform upper bound on the Logarithmic Sobolev constant $\nu_{\rho_t}$, for any $r(\theta)$ satisfying Assumption \ref{assumption_reg}: 
\begin{eqnarray}
&&\sup_{t\geq 0}\nu_{\tilde{\rho}_t} \leq 2\sqrt{\frac{2}{c_1\lambda^{-1}\Phi(0)}} +\frac{1}{c_1} + \left[2\sqrt{\frac{2}{c_1\lambda^{-1}\Phi(0)}}c_2 +\frac{2c_2}{c_1} +2\right]\inf_{r>\sqrt{\frac{c_2}{c_1\lambda^{-1}\Phi(0)}} }\frac{1+Cc_1r^4\lambda^{-1}\Phi(0)e^{2poly(r)+\gamma r^2/2}}{c_1r^2\lambda^{-1}\Phi(0)-c_2}~ \nonumber
\end{eqnarray}
where $\gamma, \Phi(0), c_1, c_2>0$ are all problem dependent constants (detailed in appendix) and $$poly(r) = C_7r^k +L_1C_1r^2 + L_1C_2r+C_8~.$$
\end{theorem}

\section{Conclusion}
We have shown that under properly chosen regularizer, the linear convergence to the global optima for noisy SGD can be obtained in training two-layer neural networks in the mean-field regime. The key step is to establish the logarithmic sobolev inequality uniformly for all Gibbs measures determined by the evolving distribution $\rho_t$. From a technical perspective, our theory implies that some growth condition on the regularizer $r(\theta)$ is sufficient for the global linear convergence, where at least cubic growth of $r(\theta)$ is required. 
Our theory also shows that the log-Sobolev constants for $\mu_t\propto e^{-U_t}$ play a crucial role in determining the convergence rate and different regularizers $r(\theta)$ produce different estimates of the log-Sobolev constants. Our current theory also has limitations and leads to some questions that worth future exploration:
\begin{itemize}
\item In practice, we usually let $r(\theta)=\beta\|\theta\|_2^2$ that grows quadratically as done in the original weight decay. Is the cubic growth condition for $r(\theta)$ a necessary condition for the global linear convergence result? To our best knowledge, there is no convergence rate proofs for the case $r(\theta)=\beta\|\theta\|_2^2$ even for a simple vanilla two-layer neural network. 
\item The main focus of this paper is on analyzing the convergence property of the continuous-time noisy SGD. In practice, we need to adopt discretization, which produces discretization errors. It leaves as our future work on how to control the discretization error carefully to keep the linear convergence property as in continuous time.
\end{itemize}

\bibliographystyle{apalike}
\bibliography{two-layer-NN}

\pagebreak\appendix
\pb

\section{Proofs of Theorem \ref{main2}} \label{A}
\begin{proof}
If we expand $U(\theta,\rho_t )$, we get
\begin{eqnarray}
&&U(\theta, \rho_t) = U(w, u; \rho_t) = \mathbb{E}_{x, y}\phi_1^\prime(f(\rho, x), y)uh(w, x) + \beta\|\theta\|_2^4~.
\end{eqnarray}
Then the gradient of $U(\theta, \rho_t)$ becomes
\begin{eqnarray}
&&\nabla U(w, u; \rho_t) = \mathbb{E}_{x, y}\left\{\phi_1^\prime(f(\rho, x), y)\begin{bmatrix}u\nabla_1 h(w, x) \\ \\  h(w, x) \end{bmatrix}\right\} + 4\beta \|\theta\|_2^2\theta ~,
\end{eqnarray}
and the Hessian of $U(\theta, \rho_t)$ is
\begin{eqnarray}
\nabla^2 U(\theta; \rho_t) = \mathbb{E}_{x, y}\left\{\phi_1^\prime(f(\rho, x), y)\begin{bmatrix}u\nabla_1^2 h(w, x) & \nabla_1 h(w, x)\\ \\ \nabla_1^T h(w, x) & 0 \end{bmatrix}\right\} + 4\beta \|\theta\|_2^2I + 8\beta\theta\theta^T~.
\end{eqnarray}
\end{proof}

Furthermore, we have
\begin{eqnarray}
&&|\lambda_{min}(\nabla^2  \mathbb{E}_{x, y}\phi_1^\prime(f(\rho, x), y)uh(w, x))|^2 \nonumber\\
&& \leq \left\|\begin{bmatrix}\mathbb{E}_{x, y}\{\phi_1^\prime(f(\rho, x), y) u\nabla_1^2 h(w, x)\} & \mathbb{E}_{x, y}\{\phi_1^\prime(f(\rho, x), y)\nabla_1 h(w, x)\}\\ \\ \mathbb{E}_{x, y}\{\phi_1^\prime(f(\rho, x), y)\nabla_1^T h(w, x)\} & 0 \end{bmatrix} \right\|_F^2 \nonumber\\
&&  = \|\mathbb{E}_{x, y}\{\phi_1^\prime(f(\rho, x), y) u\nabla_1^2 h(w, x)\}\|_F^2 + 2 \|\mathbb{E}_{x, y}\{\phi_1^\prime(f(\rho, x), y)\nabla_1 h(w, x)\}\|_2^2 \nonumber\\
&& \leq L_1^2 \|u\nabla_1^2 h(w, x)\}\|_F^2 + 2L_1^2C_3^2\nonumber\\
&& \leq dL_1^2\|\nabla_1^2 h(w, x)\}\|_{op}^2\|\theta\|_2^2 + 2L_1^2C_3^2 \nonumber\\
&& \leq dL_1^2C_4^2\|\theta\|_2^2 + 2L_1^2C_3^2 ~.
\end{eqnarray}

Hence, we have
\begin{eqnarray}
\nabla^2 U(\theta; \rho_t) \succeq \left(4\beta \|\theta\|_2^2 - \sqrt{dL_1^2C_4^2\|\theta\|_2^2 + 2L_1^2C_3^2}\right) I\succeq \left(4\beta \|\theta\|_2^2 -  \sqrt{d}L_1C_4\|\theta\|_2- \sqrt{2}L_1C_3\right) I~.
\end{eqnarray}

Obviously, we can let $$R_{m, \beta} = \frac{\sqrt{d}L_1C_4+\sqrt{dL_1^2C_4^2+16(m+\sqrt{2}\beta L_1C_3)}}{8\beta}~.$$ Then we obtain that $U(\theta; \rho_t)$ is $m$-strongly convex for $\|\theta\|_2\geq R_{m, \beta} $~. For convenience and clarity of proof, without loss of generality, we impose $h(0, x)=0$ for any $x$. Then we have $\nabla U(0, \rho_t) = 0$. Furthermore, we have the following bound showing that $U(\theta; \rho_t)$ is local Lipschitz smooth. 
\begin{eqnarray}
&& \|\nabla^2 U(\theta; \rho_t) \|_{op}\leq 12\beta \|\theta\|_2^2 +  \sqrt{dL_1^2C_4^2\|\theta\|_2^2 + 2L_1^2C_3^2}~.
\end{eqnarray}
Then, we can let 
\begin{eqnarray}
&& L_{m, \beta} = 48\beta R_{m, \beta}^2 + \sqrt{4dL_1^2C_4^2R_{m, \beta}^2 + 2L_1^2C_3^2}
\end{eqnarray}
and conclude that $U(\theta; \rho_t)$ is $L_{m, \beta}$-Lipschitz smooth for $\|\theta\|_2\leq 2R_{m, \beta} $~. Then the results can be obtained from almost the same construction and proofs of Lemma $3$ in \cite{ma2019sampling}, by modifying the global $L$-Lipschitz smooth assumption to local Lipschitz smooth assumption within the ball $B\left(0, 2R_{m, \beta}\right)$~.

\section{Proofs of Theorem \ref{main3}} \label{B}
The following lemma is essential in establishing the bound of log-Sobolev constants for distributions $ \tilde{\rho}_t(\theta)$. The proof of the lemma is nothing unusual but adapting and combining existing results in the literature. More specifically, it can be obtained by combining results from Theorem $1.4$ in \cite{bakry2008simple}, Proposition $4.3$ in \cite{cattiaux2010note}, and the Rothaus' lemma. 

\begin{lemma} \label{lemma5}
For a probability distribution $\rho_t=\exp\left(-U_t(\theta)\right)/Z$, where $U\in C^2$ and $Z = \int_{\mathbb{R}^{d+1}} e^{-U_t(\theta)}d\theta<+\infty$. If $U_t(\theta)$ satisfies the following two conditions
\begin{itemize}
\item $\nabla^2 U_t(\theta)\succeq -\Phi(\|\theta\|_2)I$ where $\Phi$ is some positive non-decreasing continuous function on $\mathbb{R}^{+}$.
\item There is some nonnegative and locally Lipschitz function $V: \mathbb{R}^{d+1}\to\mathbb{R}^{+}\cup \{0\}$ such that for some constants $c_1, c_2>0$, 
\begin{eqnarray}
&& \mathcal{L}_tV(\theta) +\|\nabla V(\theta)\|^2 \leq -c_1\|\theta\|_2^2\Phi(2\|\theta\|_2) +c_2~. \label{Lyapunov}
\end{eqnarray}
\end{itemize}
Then $\rho_t$ satisfies the log-Sobolev inequality with 
\begin{eqnarray}
&& \nu_{\rho_t} = 2\sqrt{\frac{2}{c_1\Phi(0)}} +\frac{1}{c_1} + \left[2\sqrt{\frac{2}{c_1\Phi(0)}}c_2 +\frac{2c_2}{c_1} +2\right]\inf_{r>\sqrt{\frac{c_2}{c_1\Phi(0)}} } \frac{1+C c_1\Phi(0) r^4\sup_{\|\theta\|_2\leq r}\exp(V(\theta))e^{Osc_r(U_t)}}{c_1r^2\Phi(0)-c_2}~ \nonumber
\end{eqnarray}
where $C>0$ is a universal constant and $$Osc_r(U_t) = \sup_{\|\theta\|_2\leq r}U_t(\theta)- \inf_{\|\theta\|_2\leq r}U_t(\theta)~.$$

\begin{proof}
The key step of establishing the LSI in the unbounded curvature case is by exploiting the above-tangent lemma as did in originally in \cite{barthe2008mass} and later in \cite{cattiaux2010note}. The above-tangent lemma consider a general results on a Riemannian manifold (\cite{cordero2001riemannian, cordero2006prekopa}), and we adapted it to the Euclidean space $\mathbb{R}^{d+1}$ (i.e., with zero Ricci curvature):

 let $d\mu_t(\theta) = Z^{-1}e^{-U_t(\theta)}d\theta$ be a probability measure on $\mathbb{R}^{d+1}$ and denote by $f\cdot \mu_t$ the measure with density $f$ with respect to $\mu_t$. Consider two compactly supported nonnegative functions $g$ and $h$ such that $\nu := g\cdot\mu_t$ and $ h\cdot\mu_t$ are probability measures.  Let $T(\theta) = \theta +\nabla \tau(\theta)$ be the optimal transport minimizing the quadratic transport cost and pushing forward $g\cdot\mu_t$ to $h\cdot\mu_t$. Then it holds
 \begin{eqnarray}
 &&Ent_{\mu_t}(g) \leq Ent_{\mu_t}(h) - \int_{\mathbb{R}^{d+1}}\langle \nabla \tau, \nabla g\rangle d\mu_t +  \int_{\mathbb{R}^{d+1}} \mathcal{D}_{U_t}(\theta, T(\theta))gd\mu_t~
 \end{eqnarray}
 where the convexity defect is defined as
  \begin{eqnarray}
   && \mathcal{D}_{U_t}(\theta, T(\theta)) = -\int_0^1 (1-s)\nabla \tau(\theta)^T\nabla^2 U_t(\gamma(s)) \nabla\tau(\theta)ds~,\quad \gamma(s) = \theta +s\nabla\tau(\theta)~,\quad s\in[0, 1]. \nonumber
   \end{eqnarray}

 Now, we choose a sequence of $\mu_t$-probability measures $\mu_n = h_n\mu_t$ with $h_n\in C^{\infty}_0(\mathbb{R}^{d+1})$, such that $W_2(\mu_n, \mu_t)\to 0, H(\mu_n|\mu_t) := Ent_{\mu_t} (h_n) \to 0$ and $I(\mu_n|\mu_t) := \mathcal{E}_t(\sqrt{h_n})\to 0$~. In the following, we apply the above-tangent lemma to $\nu, \mu_n$.
  
Hence, 
\begin{eqnarray}
&& \left|  \int_{\mathbb{R}^{d+1}}\langle \nabla \tau, \nabla g\rangle d\mu_t  \right| \nonumber\\
&& = \left|  \int_{\mathbb{R}^{d+1}}2\sqrt{g}\langle \nabla \tau, \nabla \sqrt{g}\rangle d\mu_t  \right|\nonumber\\
&& \leq 2 \sqrt{\int_{\mathbb{R}^{d+1}}  \|\nabla \tau\|_2^2 g d\mu_t  \int_{\mathbb{R}^{d+1}} \| \nabla \sqrt{g}\|_2^2 d\mu_t } \nonumber\\
&&  = 2 W_2(\nu, \mu_n)\sqrt{ \mathcal{E}_t(\sqrt{g})}~.
\end{eqnarray}
  
 Then we bound the defect of convexity by our condition
  \begin{eqnarray}
    &&\mathcal{D}_{U_t}(\theta, T(\theta))\nonumber\\
    &&= -\int_0^1 (1-s)\nabla \tau(\theta)^T\nabla^2 U_t(\gamma(s)) \nabla\tau(\theta)ds\nonumber\\
    &&  \leq \int_0^1 (1-s)\|\nabla\tau(\theta)\|_2^2\Phi\left(\|\gamma(s)\|_2\right)ds~.
  \end{eqnarray}
   
By the triangle inequality, we have 
\begin{eqnarray}
&&\| \nabla \tau(\theta)\|_2 =\|T(\theta)-\theta\|_2\leq 2\max\{\|T(\theta)\|_2, \|\theta\|_2\}
\end{eqnarray}
and 
\begin{eqnarray}
&&\|\gamma(s)\|_2  = \|\theta + s\nabla \tau(\theta)\|_2 = \|\theta + s(T(\theta)-\theta)\|_2  \leq 2\max\{ \|\theta\|_2, \|T(\theta)\|_2\}
\end{eqnarray}

Therefore we can bound the last term as follows:
\begin{eqnarray}
&& \int_{\mathbb{R}^{d+1}} \mathcal{D}_{U_t}(\theta, T(\theta))gd\mu_t \nonumber\\
&& \leq 2 \int_{\mathbb{R}^{d+1}}  \max\{\|T(\theta)\|_2^2, \|\theta\|_2^2\} \Phi\left( 2\max\{ \|\theta\|_2, \|T(\theta)\|_2\} \right) gd\mu_t\nonumber\\
&&\leq 2  \left(  \int_{\mathbb{R}^{d+1}} \|T(\theta)\|_2^2\Phi\left( 2  \|T(\theta)\|_2\right) gd\mu_t +  \int_{\mathbb{R}^{d+1}}  \|\theta\|_2^2 \Phi\left( 2  \|\theta\|_2 \right) gd\mu_t   \right)
\end{eqnarray}

By Lemma \ref{lemma_adjoint}, we have
\begin{eqnarray}
&& \int_{\mathbb{R}^{d+1}} \|T(\theta)\|_2^2\Phi\left( 2  \|T(\theta)\|_2\right) gd\mu_t \leq \frac{1}{c_1}(I(\mu_n|\mu_t)+c_2)  
\end{eqnarray}
and
\begin{eqnarray}
&&  \int_{\mathbb{R}^{d+1}}  \|\theta\|_2^2 \Phi\left( 2  \|\theta\|_2 \right) gd\mu_t   \leq  \frac{1}{c_1}(\mathcal{E}_t(\sqrt{g})+c_2)~.
\end{eqnarray}

Combining the above results and letting $n\to\infty$, we obtain
\begin{eqnarray}
&&   Ent_{\mu_t}(g) \leq  2 W_2(\nu, \mu)\sqrt{ \mathcal{E}_t(\sqrt{g})} +   \frac{1}{c_1}(\mathcal{E}_t(\sqrt{g})+2c_2)~.
\end{eqnarray}

Similarly, we also have
\begin{eqnarray}
&&   W_2(\nu, \mu)  \leq 2\left(\int_{\mathbb{R}^{d+1}}\|\theta\|_2^2 gd\mu_t + \int_{\mathbb{R}^{d+1}}\|\theta\|_2^2 d\mu_t  \right)\leq \frac{2}{c_1\Phi(0)}(\mathcal{E}_t(\sqrt{g})+2c_2)~.
\end{eqnarray}

Therefore, we obtain the defective LSI,
\begin{eqnarray}
&&   Ent_{\mu_t}(g) \leq  \left( 2\sqrt{\frac{2}{c_1\Phi(0)}}+\frac{1}{B}  \right)  \mathcal{E}_t(\sqrt{g}) + 2c_2\left(\sqrt{\frac{2}{c_1\Phi(0)}}+\frac{1}{B}\right)~.
\end{eqnarray}

Then, we need to estimate the spectral gap under the Lyapunov condition (\ref{Lyapunov}). 

If we let $V(\theta) = \log W(\theta)$ for some $W: \mathbb{R}^{d+1}\to [1, \infty)$, then with some simple algebra we obtain
\begin{eqnarray}
\frac{\mathcal{L}_t W(\theta)}{W(\theta)} \leq -c_1\|\theta\|_2^2\Phi(2\|\theta\|_2) +c_2\leq -c_1\|\theta\|_2^2\Phi(0) +c_2~.
\end{eqnarray}

When $\theta\in B^c(0, r)$ and choose $r\geq \sqrt{\frac{c_2}{c_2\Phi(0)}}$, we have
\begin{eqnarray}
\frac{\mathcal{L}_t W(\theta)}{W(\theta)} \leq -(c_1r^2\Phi(0) -c_2) ~.
\end{eqnarray}

When $\theta\in B(0, r)$, we have
\begin{eqnarray}
\frac{\mathcal{L}_t W(\theta)}{W(\theta)} \leq c_2~.
\end{eqnarray}

Hence, 
\begin{eqnarray}
\frac{\mathcal{L}_t W(\theta)}{W(\theta)} \leq -(c_1r^2\Phi(0) -c_2)  + c_1r^2\Phi(0) \mathbbm{1}_{B(0, r)}(\theta)~.
\end{eqnarray}

Then, 
\begin{eqnarray}
\mathcal{L}_t W(\theta)  \leq -(c_1r^2\Phi(0) -c_2)W(\theta)  + c_1r^2\Phi(0)\sup_{\|\theta\|_2\leq r} \exp\left(V(\theta)\right)\mathbbm{1}_{B(0, r)}(\theta)~.
\end{eqnarray}

By Theorem $1.4$ in \cite{bakry2008simple}, and a rough estimate of the Poincare constant for the truncated Gibbs measure by $Cr^2e^{Osc_r(U_t)}$, we obtain that $\mu_t$ also satisfies the Poincare inequality with constant 
\begin{eqnarray}
&& C_{P_t} = \inf_{r>\sqrt{\frac{c_2}{c_1\Phi(0)}} } \frac{1+C c_1\Phi(0) r^4\sup_{\|\theta\|_2\leq r}\exp(V(\theta))e^{Osc_r(U_t)}}{c_1r^2\Phi(0)-c_2}~.
\end{eqnarray}

By the Rothaus's lemma, the DLSI together with the spectral gap of $\mathcal{L}_t$ implies the LSI with constant
\begin{eqnarray}
&& \nu_{\rho_t} = 2\sqrt{\frac{2}{c_1\Phi(0)}} +\frac{1}{c_1} + \left[2\sqrt{\frac{2}{c_1\Phi(0)}}c_2 +\frac{2c_2}{c_1} +2\right]C_{P_t} \nonumber\\
&& = 2\sqrt{\frac{2}{c_1\Phi(0)}} +\frac{1}{c_1} + \left[2\sqrt{\frac{2}{c_1\Phi(0)}}c_2 +\frac{2c_2}{c_1} +2\right]\inf_{r>\sqrt{\frac{c_2}{c_1\Phi(0)}} } \frac{1+C c_1\Phi(0) r^4\sup_{\|\theta\|_2\leq r}\exp(V(\theta))e^{Osc_r(U_t)}}{c_1r^2\Phi(0)-c_2}~.\nonumber
\end{eqnarray}
\end{proof}
\end{lemma}

Now, we are ready to establish a time-dependent estimate of the Logarithmic Sobolev Constants for a family of Gibbs distributions with potential $U(\theta; \rho_t)$. 
\begin{theorem}\label{theorem6}
For $\tilde{\rho}_t(\theta) \propto \exp\left(-\lambda^{-1}U(\theta, \rho_t)\right)$, the logarithmic Sobolev Contant $\nu_{\rho_t}$ satisfies
\begin{eqnarray}
&& \nu_{\tilde{\rho}_t} = 2\sqrt{\frac{2}{c_1\lambda^{-1}\Phi(0)}} +\frac{1}{c_1} + \left[2\sqrt{\frac{2}{c_1\lambda^{-1}\Phi(0)}}c_2 +\frac{2c_2}{c_1} +2\right]\inf_{r>\sqrt{\frac{c_2}{c_1\lambda^{-1}\Phi(0)}}}\frac{1+Cc_1r^4\lambda^{-1}\Phi(0) e^{Osc_r(U_t)+\gamma r^2/2}}{c_1r^2\lambda^{-1}\Phi(0)-c_2}~ \nonumber
\end{eqnarray} 
where $\Phi(0)= \sqrt{4L_1^2C_3^2 + 2(d+1)D_4^2}$; $c_1=1$; and
\begin{eqnarray}
&& c_2 = c_2(R) :=  c_1R^2\Phi(2R) +  \gamma(d+1)+ \gamma(\gamma+L_1C_1+L_1C_3^2)R^2 +\gamma L_1C_2R +\gamma b \nonumber\\
&& =(2R)^{p/2} \sqrt{2(d+1)D_3^2} + (2R)\sqrt{2dL_1^2C_4^2} + 2(2R)^{p/4}\sqrt{(d+1)D_3D_4} \nonumber\\
&& + R^2\sqrt{4L_1^2C_3^2 + 2(d+1)D_4^2}   +  \gamma(d+1)+ \gamma(\gamma+L_1C_1+L_1C_3^2)R^2 +\gamma L_1C_2R +\gamma b \nonumber
\end{eqnarray} 
and when $p=1$, we can set any 
\begin{eqnarray}
&& R^2\geq \max\left\{1, \left(\frac{\sqrt{4L_1^2C_3^2 + 2(d+1)D_4^2}+\gamma(\gamma + L_1C_1+L_1C_3^2)}{m}\right)^{\frac{2}{p}},\right. \nonumber\\
&&\left. \left(\frac{\gamma L_1C_2}{m}\right)^{\frac{2}{1+p}}, \left(\frac{\gamma(b+d+1)}{m}\right)^{\frac{2}{2+p}},\left(\frac{2\sqrt{(d+1)D_3D_4}}{m}\right)^{\frac{4}{p}}  \right\} \nonumber
\end{eqnarray}
 and let
\begin{eqnarray}
&& \gamma = \frac{1}{m}2^{p/2}\left(\sqrt{2(d+1)} D_3+2L_1C_4\sqrt{d}\right) + 4~;
\end{eqnarray}
when $p>1$, we can set any 
\begin{eqnarray}
&& R^2\geq \max\left\{1, \left(\frac{\sqrt{4L_1^2C_3^2 + 2(d+1)D_4^2}+\gamma(\gamma + L_1C_1+L_1C_3^2)}{m}\right)^{\frac{2}{p}},\right. \nonumber\\
&&\left. \left(\frac{\gamma L_1C_2}{m}\right)^{\frac{2}{1+p}}, \left(\frac{\gamma(b+d+1)}{m}\right)^{\frac{2}{2+p}}, \left(\frac{2L_1C_4\sqrt{d}}{m}\right)^{\frac{2}{p-1}},\left(\frac{2\sqrt{(d+1)D_3D_4}}{m}\right)^{\frac{4}{p}}  \right\} \nonumber
\end{eqnarray}
 and let
\begin{eqnarray}
&& \gamma = \frac{1}{m}2^{p/2}\sqrt{2(d+1)} D_3 + 5~.
\end{eqnarray}

\begin{proof}
If we expand $U(\theta,\rho_t )$, we get
\begin{eqnarray}
&&U(\theta, \rho_t) = U(w, u; \rho_t) = \mathbb{E}_{x, y}\phi_1^\prime(f(\rho, x), y)uh(w, x) + r(\theta)~.
\end{eqnarray}
Then the gradient of $U(\theta, \rho_t)$ becomes
\begin{eqnarray}
&&\nabla U(w, u; \rho_t) = \mathbb{E}_{x, y}\left\{\phi_1^\prime(f(\rho, x), y)\begin{bmatrix}u\nabla_1 h(w, x) \\ \\  h(w, x) \end{bmatrix}\right\} + \nabla r(\theta) ~,
\end{eqnarray}
and the Hessian of $U(\theta, \rho_t)$ is
\begin{eqnarray}
&&\nabla^2 U(\theta; \rho_t) = \mathbb{E}_{x, y}\left\{\phi_1^\prime(f(\rho, x), y)\begin{bmatrix}u\nabla_1^2 h(w, x) & \nabla_1 h(w, x)\\ \\ \nabla_1^T h(w, x) & 0 \end{bmatrix}\right\} + \nabla^2 r(\theta)~.
\end{eqnarray}

Then, we have the following bound on $\nabla U(w, u; \rho_t)$,
\begin{eqnarray}
&& \|\nabla U(w, u; \rho_t)\|_2 \nonumber\\
&& = \left\|\mathbb{E}_{x, y}\left\{\phi_1^\prime(f(\rho, x), y)\begin{bmatrix}u\nabla_1 h(w, x) \\ \\  h(w, x) \end{bmatrix}\right\} + \nabla r(\theta) \right\|_2 \nonumber\\
&&\geq \left\|\nabla r(\theta) \right\|_2 -  \left\|\mathbb{E}_{x, y}\left\{\phi_1^\prime(f(\rho, x), y)\begin{bmatrix}u\nabla_1 h(w, x) \\ \\  h(w, x) \end{bmatrix}\right\} \right\|_2 \nonumber\\
&&\geq  \left(D_5\|\theta\|_2^2+D_6\right) - L_1\left( |u| \|\nabla_1 h(w, x)\|_2+ |h(w, x) |\right) \nonumber\\
&&\geq  \left(D_5\|\theta\|_2^2+D_6\right) - L_1\left( |u| C_3+ C_1\|w\|_2+C_2\right) \nonumber\\
&&\geq  \left(D_5\|\theta\|_2^2+D_6\right) - L_1\left( \|\theta\|_2 C_3+ C_1\|\theta\|_2+C_2\right)\nonumber\\
&& = D_5\|\theta\|_2^2 - (L_1C_3+C_1)\|\theta\|_2 + (D_6- L_1C_2)~.
\end{eqnarray}

Besides, we can bound the smallest eigenvalue of $\nabla^2 U(\theta; \rho_t) $ as follows,
\begin{eqnarray}
&& |\lambda_{min}(\nabla^2 U(\theta; \rho_t)|^2\nonumber\\
&& \leq trace(\nabla^2 U(\theta; \rho_t)\nabla^2 U(\theta; \rho_t)^T)\nonumber\\
&& =\underbrace{ \left\|\begin{bmatrix}\mathbb{E}_{x, y}\{\phi_1^\prime(f(\rho, x), y) u\nabla_1^2 h(w, x)\} & \mathbb{E}_{x, y}\{\phi_1^\prime(f(\rho, x), y)\nabla_1 h(w, x)\}\\ \\ \mathbb{E}_{x, y}\{\phi_1^\prime(f(\rho, x), y)\nabla_1^T h(w, x)\} & 0 \end{bmatrix} \right\|_F^2}_{T_1} \nonumber\\
&& +2 \underbrace{trace\left(\begin{bmatrix}\mathbb{E}_{x, y}\{\phi_1^\prime(f(\rho, x), y) u\nabla_1^2 h(w, x)\} & \mathbb{E}_{x, y}\{\phi_1^\prime(f(\rho, x), y)\nabla_1 h(w, x)\}\\ \\ \mathbb{E}_{x, y}\{\phi_1^\prime(f(\rho, x), y)\nabla_1^T h(w, x)\} & 0 \end{bmatrix} \nabla^2r(\theta)^T\right)}_{T_2} \nonumber\\
&& +\underbrace{ trace(\nabla^2 r(\theta)\nabla^2 r(\theta)^T)}_{T_3}~.
\end{eqnarray}

To bound $T_1$, we have
\begin{eqnarray}
&& T_1 = \|\mathbb{E}_{x, y}\{\phi_1^\prime(f(\rho, x), y) u\nabla_1^2 h(w, x)\}\|_F^2 + 2 \|\mathbb{E}_{x, y}\{\phi_1^\prime(f(\rho, x), y)\nabla_1 h(w, x)\}\|_2^2 \nonumber\\
&& \leq L_1^2 \|u\nabla_1^2 h(w, x)\}\|_F^2 + 2L_1^2C_3^2\nonumber\\
&& \leq dL_1^2\|\nabla_1^2 h(w, x)\}\|_{op}^2\|\theta\|_2^2 + 2L_1^2C_3^2 \nonumber\\
&& \leq dL_1^2C_4^2\|\theta\|_2^2 + 2L_1^2C_3^2 ~.
\end{eqnarray}
Similarly, we can bound $T_3$ as follows
\begin{eqnarray}
&& T_3 = trace(\nabla^2 r(\theta)\nabla^2 r(\theta)^T) \nonumber\\
&& \leq   (D_3\|\theta\|_2^p +D_4)^2 trace(I) \nonumber\\
&&  =  (d+1)D_3^2\|\theta\|_2^{2p} + 2(d+1)D_3D_4\|\theta\|_2^p + (d+1)D_4^2~.
\end{eqnarray}
Finally, we get the bound of $T_2$ as follows
\begin{eqnarray}
&& T_2  \leq  \left\|\begin{bmatrix}\mathbb{E}_{x, y}\{\phi_1^\prime(f(\rho, x), y) u\nabla_1^2 h(w, x)\} & \mathbb{E}_{x, y}\{\phi_1^\prime(f(\rho, x), y)\nabla_1 h(w, x)\}\\ \\ \mathbb{E}_{x, y}\{\phi_1^\prime(f(\rho, x), y)\nabla_1^T h(w, x)\} & 0 \end{bmatrix} \right\|_F\cdot \sqrt{trace(\nabla^2 r(\theta)\nabla^2 r(\theta)^T) } \nonumber\\
&& \leq \sqrt{dL_1^2C_4^2\|\theta\|_2^2 + 2L_1^2C_3^2}\cdot\sqrt{ (d+1)D_3^2\|\theta\|_2^{2p} + 2(d+1)D_3D_4\|\theta\|_2^p + (d+1)D_4^2}\nonumber\\
&& \leq 2\sqrt{dL_1^2C_4^2\|\theta\|_2^2 + 2L_1^2C_3^2}\cdot\sqrt{ (d+1)D_3^2\|\theta\|_2^{2p} + 2(d+1)D_3D_4\|\theta\|_2^p + (d+1)D_4^2} \nonumber\\
&& \leq  (d+1)D_3^2\|\theta\|_2^{2p} + 2(d+1)D_3D_4\|\theta\|_2^p + dL_1^2C_4^2\|\theta\|_2^2 + 2L_1^2C_3^2 + (d+1)D_4^2~.
\end{eqnarray}

Hence, we obtain the following bound for $|\lambda_{min}(\nabla^2 U(\theta; \rho_t)|$, 
\begin{eqnarray}
&& |\lambda_{min}(\nabla^2 U(\theta; \rho_t)|\nonumber\\
&&  \leq \sqrt{2(d+1)D_3^2\|\theta\|_2^{2p} + 4(d+1)D_3D_4\|\theta\|_2^p +2dL_1^2C_4^2\|\theta\|_2^2 + 4L_1^2C_3^2 + 2(d+1)D_4^2}\nonumber\\
&&  \leq \sqrt{2(d+1)D_3^2}\|\theta\|_2^p + \sqrt{2dL_1^2C_4^2}\|\theta\|_2 + 2\sqrt{(d+1)D_3D_4}\|\theta\|_2^{p/2} + \sqrt{4L_1^2C_3^2 + 2(d+1)D_4^2}~. \nonumber
\end{eqnarray} 
Hence, we can choose 
\begin{eqnarray}
&&\Phi(\|\theta\|_2) = \sqrt{2(d+1)D_3^2}\|\theta\|_2^p + \sqrt{2dL_1^2C_4^2}\|\theta\|_2 + 2\sqrt{(d+1)D_3D_4}\|\theta\|_2^{p/2} + \sqrt{4L_1^2C_3^2 + 2(d+1)D_4^2}~. \nonumber
\end{eqnarray}
For $\mathcal{L}_t = \nabla\cdot\nabla - \nabla U(\theta; \rho_t)\cdot \nabla,$ we now prove the following Lyapunov condition for the LSI, i.e., there is some nonnegative and locally Lipschitz function $V: \mathbb{R}^{d+1}\to\mathbb{R}^{+}\cup \{0\}$ such that
\begin{eqnarray}
&& \mathcal{L}_tV(\theta) +\|\nabla V(\theta)\|^2 \leq -c_1\|\theta\|_2^2\Phi(2\|\theta\|_2) +c_2~.
\end{eqnarray}
By setting $V(\theta) = \frac{\gamma}{2} \|\theta\|_2^2$ for some $\gamma>0$ to be chosen later, we obtain
\begin{eqnarray}
&& \mathcal{L}_tV(\theta) + \|\nabla V(\theta)\|^2= \gamma(d+1) -  \gamma \langle\nabla U(\theta; \rho_t), \theta \rangle +\gamma^2\|\theta\|_2^2~.
\end{eqnarray}

Then we have
\begin{eqnarray}
&&   \langle\nabla U(\theta; \rho_t), \theta \rangle \nonumber\\
&& =\left \langle \mathbb{E}_{x, y}\left\{\phi_1^\prime(f(\rho, x), y)\begin{bmatrix}u\nabla_1 h(w, x) \\ \\  h(w, x) \end{bmatrix}\right\} + \nabla r(\theta) , \theta \right\rangle \nonumber\\
&& \geq \left \langle \nabla r(\theta) , \theta \right\rangle  - \left|\left \langle \mathbb{E}_{x, y}\left\{\phi_1^\prime(f(\rho, x), y)\begin{bmatrix}u\nabla_1 h(w, x) \\ \\  h(w, x) \end{bmatrix}\right\} , \theta \right\rangle \right | \nonumber\\
&& \geq  m\|\theta\|_2^{2+p} -b - L_1\|\theta\|_2\sqrt{|h(w,x)|^2 + u^2\|\nabla_1h(w, x)\|_2^2}\nonumber\\
&& \geq  m\|\theta\|_2^{2+p} -b - L_1\|\theta\|_2|h(w,x)| -  L_1\|\theta\|_2|u|\|\nabla_1h(w, x)\|_2^2\nonumber\\
&& \geq  m\|\theta\|_2^{2+p} -b - L_1\|\theta\|_2(C_1\|w\|+C_2)-  L_1C_3^2\|\theta\|_2^2\nonumber\\
&& \geq m\|\theta\|_2^{2+p} - (L_1C_1+L_1C_3^2)\|\theta\|_2^2 -L_1C_2\|\theta\|_2 -b~.
\end{eqnarray}

Hence,
\begin{eqnarray}
&& \mathcal{L}_tV(\theta) + \|\nabla V(\theta)\|^2 \nonumber\\
&& \leq \gamma(d+1) -\gamma m\|\theta\|_2^{2+p} + \gamma(\gamma+L_1C_1+L_1C_3^2)\|\theta\|_2^2 +\gamma L_1C_2\|\theta\|_2 +\gamma b
\end{eqnarray} 

Now, we consider to separate the whole space $\mathbb{R}^{d+1} = B(0, R) \cup (\mathbb{R}^{d+1}\setminus B(0, R))$ where $$B(0, R) = \{\theta\in\mathbb{R}^{d+1}: \|\theta\|_2\leq R\}$$ for some $R>1$. 

When $\theta\in B(0, R)$, we have
\begin{eqnarray}
&& \mathcal{L}_tV(\theta) + \|\nabla V(\theta)\|^2 \nonumber\\
&& \leq \gamma(d+1)+ \gamma(\gamma+L_1C_1+L_1C_3^2)R^2 +\gamma L_1C_2R +\gamma b
\end{eqnarray} 
and
\begin{eqnarray}
&& -c_1\|\theta\|_2^2\Phi(2\|\theta\|_2)+ c_2 \geq  -c_1R^2\Phi(2R^2) + c_2~.
\end{eqnarray} 
Hence, without loss of generality, we can set $c_1= 1$ and
\begin{eqnarray}
&& c_2 = c_2(R) =  c_1R^2\Phi(2R) +  \gamma(d+1)+ \gamma(\gamma+L_1C_1+L_1C_3^2)R^2 +\gamma L_1C_2R +\gamma b \nonumber\\
&& =(2R)^{p/2} \sqrt{2(d+1)D_3^2} + (2R)\sqrt{2dL_1^2C_4^2} + 2(2R)^{p/4}\sqrt{(d+1)D_3D_4} \nonumber\\
&& + R^2\sqrt{4L_1^2C_3^2 + 2(d+1)D_4^2}   +  \gamma(d+1)+ \gamma(\gamma+L_1C_1+L_1C_3^2)R^2 +\gamma L_1C_2R +\gamma b
\end{eqnarray} 
Then we have
\begin{eqnarray}
&&  \mathcal{L}_tV(\theta) + \|\nabla V(\theta)\|^2 \nonumber\\
&&  \leq \gamma(d+1)+ \gamma(\gamma+L_1C_1+L_1C_3^2)R^2 +\gamma L_1C_2R +\gamma b \nonumber\\
&&  = c_2(R) - c_1R^2\Phi(2R)  \nonumber\\
&& \leq  -c_1\|\theta\|_2^2\Phi(2\|\theta\|_2)+ c_2~.
\end{eqnarray}  

When $\theta\in B^c(0, R) = \mathbb{R}^{d+1}\setminus B(0, R)$,
\begin{eqnarray}
&& \mathcal{L}_tV(\theta) + \|\nabla V(\theta)\|^2 \nonumber\\
&& \leq -\gamma m\|\theta\|_2^{2+p} + \gamma(\gamma+L_1C_1+L_1C_3^2)\|\theta\|_2^2 +\gamma L_1C_2\|\theta\|_2 + \gamma(b+d+1) ~.
\end{eqnarray} 
Now, for $p>1$, we can set any 
\begin{eqnarray}
&& R^2\geq \max\left\{1, \left(\frac{\sqrt{4L_1^2C_3^2 + 2(d+1)D_4^2}+\gamma(\gamma + L_1C_1+L_1C_3^2)}{m}\right)^{\frac{2}{p}},\right. \nonumber\\
&&\left. \left(\frac{\gamma L_1C_2}{m}\right)^{\frac{2}{1+p}}, \left(\frac{\gamma(b+d+1)}{m}\right)^{\frac{2}{2+p}}, \left(\frac{2L_1C_4\sqrt{d}}{m}\right)^{\frac{2}{p-1}},\left(\frac{2\sqrt{(d+1)D_3D_4}}{m}\right)^{\frac{4}{p}}  \right\} \nonumber
\end{eqnarray}
 and let
\begin{eqnarray}
&& \gamma = \frac{1}{m}2^{p/2}\sqrt{2(d+1)} D_3 + 5~.
\end{eqnarray}
Then we have the following bound for any $\theta\in B^c(0, R)$, 
\begin{eqnarray}
&& m\|\theta\|_2^{2+p} \nonumber\\
&&= m\|\theta\|_2^p \|\theta\|_2^2 \nonumber\\
&& \geq mR^{p}\|\theta\|_2^2 \nonumber\\
&& \geq \sqrt{4L_1^2C_3^2 + 2(d+1)D_4^2}\|\theta\|_2^2+\gamma(\gamma + L_1C_1+L_1C_3^2)\|\theta\|_2^2~.
\end{eqnarray}
Similarly, we also have the following bound for any $\theta\in B^c(0, R)$,
\begin{eqnarray}
&& m\|\theta\|_2^{2+p}\geq 2\sqrt{d}L_1C_4\|\theta\|_2^3, \\
&& m\|\theta\|_2^{2+p}\geq  2\sqrt{(d+1)D_3D_4} \|\theta\|_2^{p/2+2},\\
&& m\|\theta\|_2^{2+p}\geq \gamma L_1C_2\|\theta\|_2, \\
&& m\|\theta\|_2^{2+p}\geq  \gamma(b+d+1)~.
\end{eqnarray}
Thus,
\begin{eqnarray}
&& \mathcal{L}_tV(\theta) + \|\nabla V(\theta)\|^2 \nonumber\\
&& \leq -\gamma m\|\theta\|_2^{2+p} + \gamma(\gamma+L_1C_1+L_1C_3^2)\|\theta\|_2^2 +\gamma L_1C_2\|\theta\|_2 + \gamma(b+d+1) \nonumber\\
&&=-2^{p/2}\sqrt{2(d+1)} D_3\|\theta\|_2^{p+2}  -5m\|\theta\|_2^{2+p} + \gamma(\gamma+L_1C_1+L_1C_3^2)\|\theta\|_2^2 +\gamma L_1C_2\|\theta\|_2 + \gamma(b+d+1)\nonumber\\
&&= -2^{p/2}\sqrt{2(d+1)} D_3\|\theta\|_2^{p+2} - 2\sqrt{d}L_1C_4\|\theta\|_2^3- 2\sqrt{(d+1)D_3D_4} \|\theta\|_2^{p/2+2}-  \sqrt{4L_1^2C_3^2 + 2(d+1)D_4^2}\|\theta\|_2^2 \nonumber\\
&& = -\|\theta\|_2^2\Phi(2\|\theta\|_2)~.
\end{eqnarray}

Similarly, for $p=1$, we can set any 
\begin{eqnarray}
&& R^2\geq \max\left\{1, \left(\frac{\sqrt{4L_1^2C_3^2 + 2(d+1)D_4^2}+\gamma(\gamma + L_1C_1+L_1C_3^2)}{m}\right)^{\frac{2}{p}},\right. \nonumber\\
&&\left. \left(\frac{\gamma L_1C_2}{m}\right)^{\frac{2}{1+p}}, \left(\frac{\gamma(b+d+1)}{m}\right)^{\frac{2}{2+p}},\left(\frac{2\sqrt{(d+1)D_3D_4}}{m}\right)^{\frac{4}{p}}  \right\} \nonumber
\end{eqnarray}
 and let
\begin{eqnarray}
&& \gamma = \frac{1}{m}2^{p/2}\left(\sqrt{2(d+1)} D_3+2L_1C_4\sqrt{d}\right) + 4~.
\end{eqnarray}

Then we can also get,
\begin{eqnarray}
&& \mathcal{L}_tV(\theta) + \|\nabla V(\theta)\|^2 \nonumber\\
&& \leq -\gamma m\|\theta\|_2^{2+p} + \gamma(\gamma+L_1C_1+L_1C_3^2)\|\theta\|_2^2 +\gamma L_1C_2\|\theta\|_2 + \gamma(b+d+1) \nonumber\\
&&=-2^{p/2}(\sqrt{2(d+1)} D_3+2L_1C_4\sqrt{d})\|\theta\|_2^{p+2}  -5m\|\theta\|_2^{2+p}  \nonumber\\
&&+ \gamma(\gamma+L_1C_1+L_1C_3^2)\|\theta\|_2^2 +\gamma L_1C_2\|\theta\|_2 + \gamma(b+d+1)\nonumber\\
&&=-2^{p/2}(\sqrt{2(d+1)} D_3+2L_1C_4\sqrt{d})\|\theta\|_2^{p+2} - 2\sqrt{(d+1)D_3D_4} \|\theta\|_2^{p/2+2}-  \sqrt{4L_1^2C_3^2 + 2(d+1)D_4^2}\|\theta\|_2^2 \nonumber\\
&& = -\|\theta\|_2^2\Phi(2\|\theta\|_2)~.
\end{eqnarray} 

In summary, for any $p\geq 1$, and $\theta \in\mathbb{R}^d$, if we set $V(\theta) = \frac{\gamma}{2}\|\theta\|_2^2$,we have
\begin{eqnarray}
&& \mathcal{L}_tV(\theta) + \|\nabla V(\theta)\|^2  \leq  -\|\theta\|_2^2\Phi(2\|\theta\|_2) + c_2~.
\end{eqnarray} 
where 
\begin{eqnarray}
&& c_2 = c_2(R) :=  c_1R^2\Phi(2R) +  \gamma(d+1)+ \gamma(\gamma+L_1C_1+L_1C_3^2)R^2 +\gamma L_1C_2R +\gamma b \nonumber\\
&& =(2R)^{p/2} \sqrt{2(d+1)D_3^2} + (2R)\sqrt{2dL_1^2C_4^2} + 2(2R)^{p/4}\sqrt{(d+1)D_3D_4} \nonumber\\
&& + R^2\sqrt{4L_1^2C_3^2 + 2(d+1)D_4^2}   +  \gamma(d+1)+ \gamma(\gamma+L_1C_1+L_1C_3^2)R^2 +\gamma L_1C_2R +\gamma b \nonumber
\end{eqnarray} 
and when $p=1$, we can set any 
\begin{eqnarray}
&& R^2\geq \max\left\{1, \left(\frac{\sqrt{4L_1^2C_3^2 + 2(d+1)D_4^2}+\gamma(\gamma + L_1C_1+L_1C_3^2)}{m}\right)^{\frac{2}{p}},\right. \nonumber\\
&&\left. \left(\frac{\gamma L_1C_2}{m}\right)^{\frac{2}{1+p}}, \left(\frac{\gamma(b+d+1)}{m}\right)^{\frac{2}{2+p}},\left(\frac{2\sqrt{(d+1)D_3D_4}}{m}\right)^{\frac{4}{p}}  \right\} \nonumber
\end{eqnarray}
 and let
\begin{eqnarray}
&& \gamma = \frac{1}{m}2^{p/2}\left(\sqrt{2(d+1)} D_3+2L_1C_4\sqrt{d}\right) + 4~;
\end{eqnarray}
when $p>1$, we can set any 
\begin{eqnarray}
&& R^2\geq \max\left\{1, \left(\frac{\sqrt{4L_1^2C_3^2 + 2(d+1)D_4^2}+\gamma(\gamma + L_1C_1+L_1C_3^2)}{m}\right)^{\frac{2}{p}},\right. \nonumber\\
&&\left. \left(\frac{\gamma L_1C_2}{m}\right)^{\frac{2}{1+p}}, \left(\frac{\gamma(b+d+1)}{m}\right)^{\frac{2}{2+p}}, \left(\frac{2L_1C_4\sqrt{d}}{m}\right)^{\frac{2}{p-1}},\left(\frac{2\sqrt{(d+1)D_3D_4}}{m}\right)^{\frac{4}{p}}  \right\} \nonumber
\end{eqnarray}
 and let
\begin{eqnarray}
&& \gamma = \frac{1}{m}2^{p/2}\sqrt{2(d+1)} D_3 + 5~.
\end{eqnarray}
\end{proof}
\end{theorem}

Now, for $t\geq 0$, we are ready to prove Theorem \ref{main3}, obtaining the following uniform bound for $\nu_{\rho_t}$.
\begin{theorem} 
We have the following uniform bound on the Logarithmic Sobolev constant $\nu_{\rho_t}$.
\begin{eqnarray}
&& \sup_{t\geq 0}\nu_{\tilde{\rho}_t} \leq 2\sqrt{\frac{2}{c_1\lambda^{-1}\Phi(0)}} +\frac{1}{c_1} \nonumber\\
&&+ \left[2\sqrt{\frac{2}{c_1\lambda^{-1}\Phi(0)}}c_2 +\frac{2c_2}{c_1} +2\right]\inf_{r>\sqrt{\frac{c_2}{c_1\lambda^{-1}\Phi(0)}} }\frac{1+Cc_1r^4\lambda^{-1}\Phi(0)e^{2poly(r)+\gamma r^2/2}}{c_1r^2\lambda^{-1}\Phi(0)-c_2}:=\nu ~ \nonumber
\end{eqnarray}
where $poly(r) = C_7r^k +L_1C_1r^2 + L_1C_2r+C_8$.

\begin{proof}
For any $t\geq 0$, we have
\begin{eqnarray}
&&|U(\rho_t; \theta)|  \nonumber\\
&&= |  \mathbb{E}_{x, y}\phi_1^\prime(f(\rho, x), y)\tilde{h}(\theta, x) + r(\theta) | \nonumber\\
&& \leq L_1\|\theta\|_2(C_1\|\theta\|_2+C_2) +C_7\|\theta\|_2^k +C_8  \nonumber\\
&& = C_7\|\theta\|_2^k +L_1C_1\|\theta\|_2^2 + L_1C_2\|\theta\|_2+C_8 
\end{eqnarray}
Hence, we obtain
\begin{eqnarray}
&&\sup_{\|\theta\|_2\leq r}|U(\rho_t; \theta)|  \leq C_7r^k +L_1C_1r^2 + L_1C_2r+C_8 := poly(r) 
\end{eqnarray}
and 
\begin{eqnarray}
&&Osc_r(U_t) \leq 2\sup_{\|\theta\|_2\leq r}|U(\rho_t; \theta)|  \leq 2(C_7r^k +L_1C_1r^2 + L_1C_2r+C_8) = 2poly(r)~.
\end{eqnarray}
\end{proof}
\end{theorem}

The following property of the operator $\mathcal{L}_t$ is useful in establishing the DLSI in our analysis. It appears in many literatures about LSI, such as \cite{cattiaux2009lyapunov} and \cite{cattiaux2010note}, but we include the proof here for completeness. 

\begin{lemma}\label{lemma_adjoint}
Let $V$ be a nonnegative and locally Lipschitz function such that $\mathcal{L}_tV + \|\nabla V\|_2^2 \leq -\phi$, where $\phi$ is lower bounded. Then
\begin{eqnarray}
\int \phi g^2d\mu_t \leq \mathcal{E}_t(g), \quad g\in  \mathbb{D}_2(\mathcal{L}_t).
\end{eqnarray}
\begin{proof}
Since $\phi \wedge N$ also satisfies the condition and can approach $\phi$ by letting $N\to\infty$, we can assume that $\phi$ is bounded. Furthermore, since $\mathbb{D}_2(\mathcal{L}_t)$ is a dense subset of $L^2(\mu_t)$, we can approach any $g\in \mathbb{D}_2(\mathcal{L}_t)$ by $(g_n)\subset C_0^\infty(\mathbb{R}^{d+1}): \|g_n-g\|_{L^2(\mu_t)}^2 + \mathcal{E}_t(g_n - g)\to 0$. Hence it is sufficient to prove for $g\in C_0^\infty(\mathbb{R}^{d+1})$. By the self adjoint property, we have
\begin{eqnarray}
\langle -\mathcal{L}_t V, g^2 \rangle_{L^2(\mu_t)} = \langle  V, -\mathcal{L}_t g^2 \rangle_{L^2(\mu_t)} =  \langle  \nabla V, \nabla (g^2) \rangle_{L^2(\mu_t)} = \int_{\mathbb{R}^{d+1}} \nabla V\cdot \nabla(g^2)d\mu_t~.
\end{eqnarray}

Therefore, we have
\begin{eqnarray}
&& \int_{\mathbb{R}^{d+1}} \phi g^2 d\mu_t \nonumber\\
&& \leq \int_{\mathbb{R}^{d+1}} (-\mathcal{L}_tV - \|\nabla V\|_2^2) g^2 d\mu_t \nonumber\\
&& = \int_{\mathbb{R}^{d+1}} (\nabla V\cdot \nabla(g^2) - \|\nabla V\|_2^2g^2)  d\mu_t\nonumber\\
&& = \int_{\mathbb{R}^{d+1}} (2g\nabla V\cdot \nabla g - \|\nabla V\|_2^2g^2 ) d\mu_t\nonumber\\
&& \leq \int_{\mathbb{R}^{d+1}} \|\nabla g\|_2^2 d\mu_t = \mathcal{E}_t(g)~
\end{eqnarray}
where in the last inequality, we use $2g\nabla V\cdot \nabla g \leq \|\nabla V\|_2^2g^2 +  \|\nabla g\|_2^2 $.
\end{proof}
\end{lemma}

\section{Experimental Results} \label{experiments}
We conduct experiments to demonstrate the role of different regularizers to the convergence.

\begin{figure}[H]
    \centering
    \includegraphics[width=10cm]{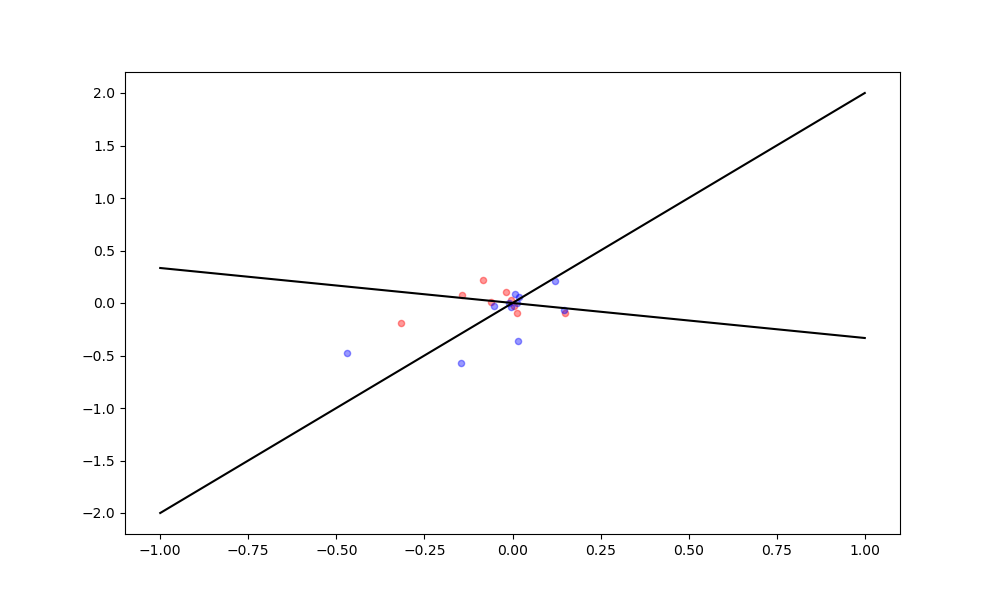}
    \caption{Initialized neurons can be visualized in a two-dimensional plane by $uw\in\mathbb{R}^2$. Positive wighted neurons ($u\geq 0$) are in red and negative weighted neurons ($u<0$) are in blue. Two black lines are the two hidden neurons of the teacher network.}
    \label{fig:untrained}
\end{figure}

In the experiment, we consider $d=2$ and the data is generated from a two-layer teacher neural network with two hidden neurons $w_1 = [1, 2]^T, u_1 = 1.1, w_2 = [-3, 1]^T, u_2=-3.2$. We choose smoothed-ReLU as activations according to our theoretic results. 
We consider an overparametrized student network with $20$ hidden neurons.  The weights are initialized by standard Gaussian. Initialized weights are shown in \autoref{fig:untrained}.
We choose the regularized square loss, with $\beta=1.0$ and $\lambda=1.0$. The initial learning rate is $0.0001$ and decays every $100$ epochs. We trained the student network for $200$ epochs and we generate $200$ data points from the teacher network to train the student network in each epoch. 

We train the two student networks with the same initialized weights but different regularizers: $r_1(\theta) = \frac{\beta}{2}\|\theta\|_2^2$ for one network and $r_2(\theta)=\frac{\beta}{3}\|\theta\|_2^3$ for another network. 

\begin{figure}[H]
    \centering
    \includegraphics[width=10cm]{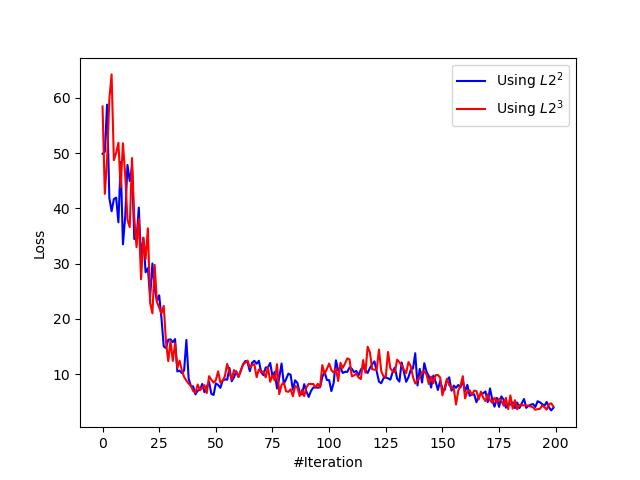}
    \caption{Square Loss}
    \label{fig:loss}
\end{figure}

\begin{figure}[H]
    \centering
    \includegraphics[width=10cm]{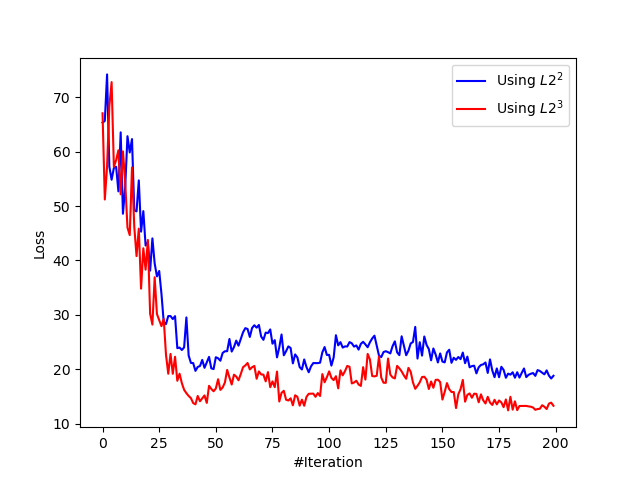}
    \caption{Regularized Square Loss}
    \label{fig:reg_loss}
\end{figure}

\begin{figure}[H]
    \centering
    \includegraphics[width=10cm]{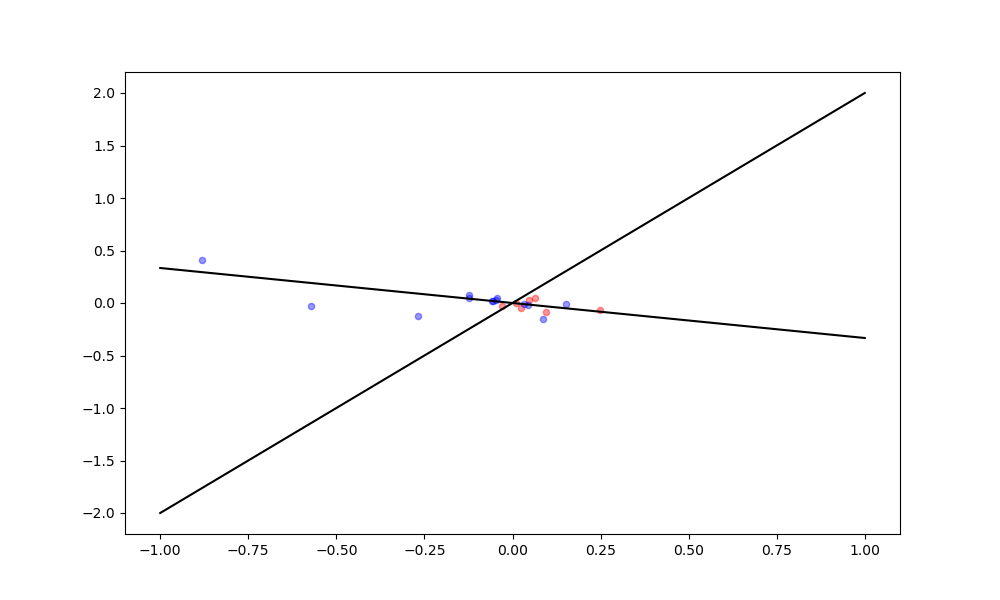}
    \caption{Learned Weights for $r_1(\theta) = \frac{\beta}{2}\|\theta\|_2^2$}
    \label{fig:loss_2}
\end{figure}

\begin{figure}[H]
    \centering
    \includegraphics[width=10cm]{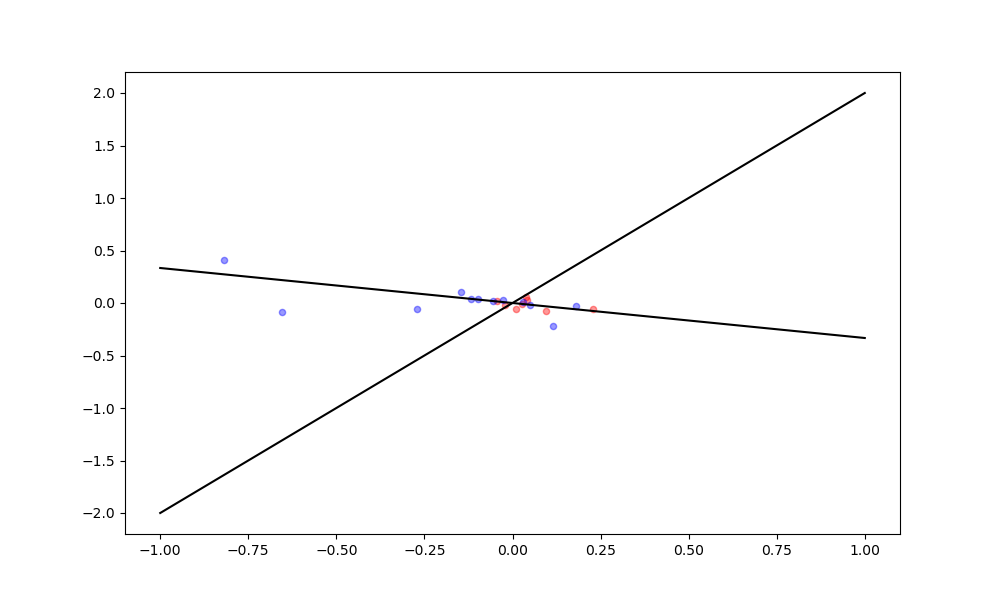}
    \caption{Learned Weights for $r_2(\theta)=\frac{\beta}{3}\|\theta\|_2^3$}
    \label{fig:loss_3}
\end{figure}

\autoref{fig:loss} shows the evolution of unregularized square loss and \autoref{fig:reg_loss} shows the regularized square loss of two student networks during the training. The learned weights for two networks are shown in \autoref{fig:loss_2} and \autoref{fig:loss_3} respectively.

From the experiments, we know that the cubic regularizer $r_2(\theta)$ has similar convergence  to $r_1(\theta)$ for both  the regularized objective as shown in \autoref{fig:reg_loss} and the unregularized objective as shown in \autoref{fig:loss}. Furthermore, the results demonstrate that both regularizers can learn weights and positions efficiently in the mean-field regime while the NTK regime suffers from lazy training and can only learn the weight $u$ of hidden neurons. The experiments show that the cubic regularizer achieves comparable performance to the original weight decay, with additional theoretic linear convergence guarantees in the many-particle and continuous time limit.

\section{Further Discussions} \label{discussion}
\subsection{Intuitions behind Theorem \ref{main2} and Theorem \ref{main3}}
As shown in our analysis, establishing the uniform LSI for  $\tilde{\rho}_t \propto \exp(-\lambda^{-1}U(\theta, \rho_t))$ along the trajectory is the key technical contribution to obtain the linear convergence of two-layer NN trained by noisy SGD in the mean-field regime. This is purely a math problem and we need to adopt the literature of LSI, where a general principle is to control the curvature lower bound of $U(\theta, \rho_t)$ to obtain the LSI. 

The most fundamental idea to establish the LSI is by the Baker-Emery criteria, when the potential $U(\theta, \rho_t))$ is a strongly convex function of $\theta$ (i.e., with positive curvature lower bound). However, for neural networks, the convexity of potential $U(\theta, \rho_t))$ is hard to satisfy. One can generalize the Bakery-Emery criteria to the nonconvex potentials by the Holley-Stroock perturbation lemma: if the nonconvex potential is a bounded perturbation of some strongly convex potential, then it also satisfies the LSI. 

This is exactly what [Nitanda et al. '22] \& [Chizat '22] adopts to establish LSI: they assume that the neural networks is a bounded perturbation of the $\|\theta\|_
2^2$ regularizer. From a technical perspective, treating the neural network as the perturbation of the regularizer seems not satisfactory. In this sense, the convergence analysis is a result from the regularizer. The bounded assumption is not satisfied by the vanilla two-layer network unless we fix the parameter of the second layer. 

Hence, Theorem \ref{main2} in this paper can be considered as a nontrivial extension to the result of [Nitanda et al. '22] \& [Chizat '22], where we consider the regularizer with growth  at least $\|\theta\|_2^4$. In such sense, there is more ``room'' for nonconvex perturbation and the ``room'' can fit the vanilla two-layer neural networks.  As shown in our derivations, we see that the Hessian of the nonconvex potential contributed by neural networks grows with $\mathcal{O}(-\|\theta\|_2I)$ whereas the Hessian of the regularizer $\|\theta\|_2^p$ grow with $\mathcal{O}(\|\theta\|_2^{p-2}I)$. Hence as long as $p-2>1$, that is $p\geq 4$, we can always ensure the Hessian of the nonconvex potential be positive outside a ball in $\mathbb{R}^{d+1}$.

Finally in Theorem \ref{main3}, we aim to find a tightest possible ``room'' that can fit vanilla two-layer neural networks. Therefore, we adopt the recent state-of-art results in the LSI literature to estimate the Log-Sobolev constant for vanilla two-layer neural networks. In particular, Cattiaux et al. (2010) obtains that when the lower bound on the Hessian of the potential diverges to $-\infty$, but can be controlled by some nondecreasing function $\Phi(\theta)$, then the LSI can also be obtained under the Lyapunov conditions in (33). 

For the behavior of the right-hand side of the inequality in two theorems, one could observe that there exists an optimal $\lambda$ for the entropy regularizer that produce the sharpest bound. One could use this optimal $\lambda$ in model selection, where we can choose the most suitable noise level in noisy SGD to accelerate the convergence. Another take-home message is that smoother loss and activations guarantee better convergence, which is also desirable intuitively.

\subsection{Is ~$p\geq 3$~ for ~$r(\theta) = \beta \|\theta\|_2^p$~ necessary?}
After some careful analysis on the potential contributed by two-layer neural networks and Lyapunov conditions in equation (33), we can easily found that the best choice of $\Phi(\theta)$ is of order $\|\theta\|_2$ and the weakest regularization is of order $ \|\theta\|_2^3$ to establish the Lyapunov condition in (33). 

If we want to relax the restrictions on $p\geq 3$, we need to develop new mathematical tools to establish LSI better than Cattiaux et al. (2010), which should be the job for mathematicians.


\end{document}